\pdfoutput=1

\documentclass[letter, 10 pt, journal, twoside]{IEEEtran}

\usepackage[colorlinks,bookmarksopen,bookmarksnumbered,citecolor=red,urlcolor=red]{hyperref}

\newcommand\BibTeX{{\rmfamily B\kern-.05em \textsc{i\kern-.025em b}\kern-.08em
T\kern-.1667em\lower.7ex\hbox{E}\kern-.125emX}}

\usepackage{graphics} 
\usepackage[table,dvipsnames]{xcolor}
\usepackage{amsmath, amssymb} 
\usepackage{cases} 
\usepackage{siunitx}
\usepackage{placeins}
\usepackage{afterpage}
\usepackage{booktabs} 
\newcommand{\ra}[1]{\renewcommand{\arraystretch}{#1}} 
\usepackage{tabstackengine}
\stackMath
\usepackage[shortcuts]{extdash}
\usepackage{textcomp}
\usepackage{array}
\usepackage{multicol, multirow, rotating}
\usepackage[nolist]{acronym}
\usepackage{verbatim}
\usepackage{enumitem}
\usepackage{textcomp} 
\usepackage{blindtext}

\usepackage{tikz}
\usetikzlibrary{arrows}

\usepackage[edges]{forest}
\useforestlibrary{linguistics}

\definecolor{folderbg}{RGB}{124,166,198}
\definecolor{folderborder}{RGB}{110,144,169}
\newlength\Size
\tikzset{%
  folder/.pic={%
    \filldraw [draw=folderborder, top color=folderbg!50, bottom color=folderbg] (-1.05*\Size,0.2\Size+5pt) rectangle ++(.75*\Size,-0.2\Size-5pt);
    \filldraw [draw=folderborder, top color=folderbg!50, bottom color=folderbg] (-1.15*\Size,-\Size) rectangle (1.15*\Size,\Size);
  },
  file/.pic={%
    \filldraw [draw=folderborder, top color=folderbg!5, bottom color=folderbg!10] (-\Size,.4*\Size+5pt) coordinate (a) |- (\Size,-1.2*\Size) coordinate (b) -- ++(0,1.6*\Size) coordinate (c) -- ++(-5pt,5pt) coordinate (d) -- cycle (d) |- (c) ;
  },
}

\forestset{%
  declare autowrapped toks={pic me}{},
  declare boolean register={pic root},
  pic dir tree/.style={%
    for tree={%
      folder,
      font=\ttfamily,
      grow'=0,
    },
    before typesetting nodes={%
      for tree={%
        edge label+/.option={pic me},
      },
      if pic root={
        tikz+={
          \pic at ([xshift=\Size].west) {folder};
        },
        align={l}
      }{},
    },
  },
  pic me set/.code n args=2{%
    \forestset{%
      #1/.style={%
        inner xsep=2\Size,
        pic me={pic {#2}},
      }
    }
  },
  pic me set={directory}{folder},
  pic me set={file}{file},
}
\usepackage{custom/chb_commands}

\graphicspath{{./}{figures/}}

\newcommand{\StereoCamAngle}{60}
\begin{document}
\markboth{}{Brommer \MakeLowercase{\textit{et al.}}: INSANE: Cross-Domain UAV Datasets} 
\title{INSANE: Cross-Domain UAV Datasets with Increased Number of Sensors for developing Advanced and Novel Estimators}
\begin{acronym}
\acro{RPM}[RPM]{Revolutions per Minute}
\acro{MAV}[MAV]{Micro Aerial Vehicle}
\acro{UAV}[UAV]{Unmanned Aerial Vehicle}
\acro{GNSS}[GNSS]{Global Navigation Satellite System}
\acro{RTK}[RTK]{Real-Time Kinematic}
\acro{IMU}[IMU]{Inertial Measurement Unit}
\acro{LRF}[LRF]{Laser Range Finder}
\acro{UWB}[UWB]{Ultra-Wide-Band}
\acro{FoV}[FoV]{Field of View}
\acro{VIO}[VIO]{Visual Inertial Odometry}
\acro{EMI}[EMI]{Electromagnetic Interference}
\acro{MEMS}[MEMS]{Micro-Electromechanical Systems}
\acro{SoC}[SoC]{System on Chip}
\acro{FoV}[FoV]{Field of View}
\acro{UHF}[UHF]{Ultra High Frequency}
\acro{SfM}[SfM]{Structure-from-Motion}
\acro{GPIO}[GPIO]{General Purpose Input/Output}
\end{acronym}

\author{Christian Brommer$^{1}$, Alessandro Fornasier$^{1}$, Martin Scheiber$^{1}$, Jeff Delaune$^{2}$, Roland Brockers$^{2}$, Jan Steinbrener$^{1}$ and Stephan Weiss$^{1}$
\thanks{$^{1}$Control of Networked Systems Group of the University of Klagenfurt, Austria. E-Mail:
        {\tt\footnotesize { \{ }}
        {\tt\footnotesize {\href{mailto:christian.brommer@ieee.org}{christian.brommer}}},
        {\tt\footnotesize {\href{mailto:alessandro.fornasier@aau.at}{alessandro.fornasier}}},
        {\tt\footnotesize {\href{mailto:martin.scheiber@aau.at}{martin.scheiber}}},
        {\tt\footnotesize {\href{mailto:jan.steinbrener@ieee.org}{jan.steinbrener}}},
        {\tt\footnotesize {\href{mailto:stephan.weiss@ieee.org}{stephan.weiss}}}
        {\tt\footnotesize { \}@ieee.org}}
        }%
\thanks{$^{2}$Jet Propulsion Laboratory, California Institute of Technology. E-Mail:
        {\tt\footnotesize {\href{mailto:roland.brockers@jpl.nasa.gov}{roland.brockers@jpl.nasa.gov}}},
        {\tt\footnotesize {\href{mailto:jeff.h.delaune@jpl.nasa.gov}{jeff.h.delaune@jpl.nasa.gov}}}
        }%

\thanks{\noindent
\noindent
``The INSANE dataset: Large number of sensors for challenging UAV flights in Mars analog, outdoor, and out-/indoor transition scenarios,''
Published in
\textit{The International Journal of Robotics Research}, 
\textbf{43}(8), pp.~1083--1113. 
\copyright~2024 The Author(s). 
DOI: \href{https://doi.org/10.1177/02783649241227245}{10.1177/02783649241227245}.
}
}


\maketitle
\begin{abstract}
For real-world applications, autonomous mobile robotic platforms must be capable of navigating safely in a multitude of different and dynamic environments with accurate and robust localization being a key prerequisite. 
To support further research in this domain, we present the INSANE datasets (Increased Number of Sensors for developing Advanced and Novel Estimators) - a collection of versatile \ac{MAV} datasets for cross-environment localization.
The datasets provide various scenarios with multiple stages of difficulty for localization methods. These scenarios range from trajectories in the controlled environment of an indoor motion capture facility, to experiments where the vehicle performs an outdoor maneuver and transitions into a building, requiring changes of sensor modalities, up to purely outdoor flight maneuvers in a challenging Mars analog environment to simulate scenarios which current and future Mars helicopters would need to perform.
The presented work aims to provide data that reflects real-world scenarios and sensor effects. The extensive sensor suite includes various sensor categories, including multiple \acp{IMU} and cameras.
Sensor data is made available as unprocessed measurements and each dataset provides highly accurate ground~truth, including the outdoor experiments where a dual \ac{RTK} \ac{GNSS} setup provides sub-degree and centimeter accuracy (1-sigma).
The sensor suite also includes a dedicated high-rate \ac{IMU} to capture all the vibration dynamics of the vehicle during flight to support research on novel machine learning-based sensor signal enhancement methods for improved localization.
The datasets and post-processing tools are available at: \url{https://sst.aau.at/cns/datasets/insane-dataset/}
\end{abstract}


\section{Introduction}
Real-world datasets are an essential part of the research and development process in the field of robotics.
When developing new methods and algorithms, one of the first steps is to test and prove an approach with flawless simulated data sequences, followed by more advanced verification in which the simulated data needs to reflect real-world sensor behaviors such as noise, non-Gaussian signal distributions, and environment-based signal degradation.
Modeling realistic sensor signals and their degradation linked to the environment is a difficult, if not impossible, task.
At this stage, real-world datasets with accurate ground~truth are a prerequisite to move from the ivory tower to real-world applications.

Existing \ac{UAV} datasets focus on isolated topics such as \ac{VIO}, indoor navigation, or vehicle control with aspects to energy efficiency as presented by \cite{Rodrigues2021}.
Datasets focusing on outdoor \ac{UAV} applications are sparse, and the provided ground~truth for subsequent algorithm development and evaluation is not of sufficient quality. 
In addition, research towards multi-environment \ac{UAV} operations is progressively increasing. One example is the transition of \acp{UAV} from outdoor environments to indoor locations and vice versa.
Such operations cause changes to the sensor availability, such as \ac{GNSS} sensors, which become unavailable in particular phases. Other side effects include anomalies in the magnetic field, close to building structures, which affect the readings of a magnetometer, or changing light conditions affecting \ac{VIO} approaches.
It may also require changes of the navigation reference frames if the available sensors provide relative navigation, e.g., \ac{VIO}.

Autonomous tasks requiring such scenarios are package delivery applications, automated emergency response, vessel inspection 
(e.g., the European BugWright2\footnote{\url{https://www.bugwright2.eu/project/}} project),
and long-term environmental surveying for, e.g., agricultural applications \cite{Malyuta2020}.
Corresponding research for transitioning robotic vehicles includes \cite{Congram2021} and long-duration \ac{UAV} autonomy with possible indoor recharging \cite{Brommer2018_lda}.

Unfortunately, openly available multi-environment data, which is necessary to elevate this field of research and to move from simulated scenarios into the real world, especially for \acp{UAV}, does not exist.
Simulated environments include the work introduced by \cite{Fornasier2021_vinseval} and \cite{Wang2020} as well as datasets with real sensor data but artificially augmented vision presented by \cite{Antonini2020_ijrrblackbird}.
Yet, using simulated data for this aspect is only suitable for initial development stages as it does not account for realistic environmental effects introduced to a sensor, such as near infrastructure affecting \ac{GNSS} or \ac{UWB} signals during an indoor-outdoor transition.

While the applications mentioned before require a versatile multi-sensor setup to complete a task and gain sufficient knowledge about the environment, we also want to mention the possible aspects for which this dataset can aid the development and improvement of novel multi-sensor fusion approaches.
Figure~\ref{fig:dataset_num_comp} in the related work section shows the increased number of sensors and a wide variety of sensor modalities of the INSANE dataset compared to state-of-the-art datasets. Other UAV datasets do not carry a high number of sensors due to restricted payloads.

The presented INSANE flight dataset aims to overcome this limitation.
With a total of 18 sensors, ranging from \mbox{high-resolution} navigation images over \mbox{high-rate} and \mbox{multi-IMU} signals to \mbox{multi-\ac{GNSS}} and \ac{UWB} data.
This dataset provides the opportunity for the validation of localization approaches with aspects to centralized estimation and modular integration of sensor information, scalability of new methodologies, and properties to robustness and sensor switching approaches under real-world conditions.
It further promotes the development for robust mapping and perception as well as the comparison of learning vs. classical approaches.
Research on these topics will benefit from the increased number of sensors, the broader modality of sensor types, and the fact that this sensor data is subject to real-world sensor degradation.
In addition to the variety of sensors, the dataset also focuses on scenarios for cross-environment robotics by providing a variety of cross-domain and multi-environment flight datasets with highly accurate ground~truth for position and orientation (6~DoF) for indoor and outdoor setups.

For the generation of ground~truth data, special attention and effort was given to the acquisition of high-quality raw (unprocessed) measurements provided by the sensors that are required to generate this ground~truth. The ground~truth provided for the presented dataset is given as an absolute entity and is not expressed relative to existing localization algorithms, as done by distinct related work. Thus, comparing localization algorithms to the ground~truth provided by this work allows for a definite evaluation of errors without restrictions to specific metrics.

The same platform was used to record 27~datasets with accumulated trajectories of more than \SI{2}{\kilo\meter} while operating in four distinct environmental domains.
It provides the necessary data to validate individual algorithm setups in a controlled environment and gradually increases the difficulty for successive proof of algorithms and methods.

\noindent The main features are:
\begin{itemize}[leftmargin=*,noitemsep,topsep=0pt,parsep=0pt,partopsep=0pt]
    \item 6~DoF absolute ground~truth with centimeter and sub-degree accuracy (1-sigma) for outdoor datasets.
    \item Indoor trajectories with motion capture ground~truth (6~DoF millimeter and sub-degree accuracy) for the initial proof of algorithms.
    \item Outdoor to indoor transition trajectories with continuous absolute ground~truth.
    \item Trajectories in a Mars analog desert environment for Mars-Helicopter analog setups, including various ground structures, cliff flight over, and cliff-wall traversing trajectories for mapping.
    \item Vehicle and sensor integrity, including intrinsic information such as static \ac{IMU} data and \ac{RPM} correlated vibration data.
    \item Real-world sensor effects and degradation posed by individual scenarios.
    \item Initialization sequences for \ac{VIO} algorithms.
    \item Inter-sensor calibrations in pre-calculated form and unprocessed calibration data sequences for custom calibration routines.
\end{itemize}

\subsection{Structure}

We first discuss related work and discuss differences between existing datasets in Section~\ref{sec:related_work}.
Section~\ref{sec:system_setup} discusses the full system setup, outlining the properties of the vehicle in~\ref{sec:vehicle_config} and individual aspects of all sensors, including the general module synchronization approach in~\ref{sec:sensors} as well as a test bench setup for a dedicated vibration test in~\ref{sec:vibration_test_bench}.

Section~\ref{sec:environment_and_experiments} will discuss the individual environmental domains and challenges for the data that was recorded in these environments.
These include simple and controlled indoor environments, more involved outdoor and outdoor/indoor transition setups at a semi-urban university location, and the Mars analog desert environment, which utilizes the full sensor suite and can be interpreted as an \mbox{off-world} setup and that also provides challenging VIO scenarios with ground structures that are semi-homogeneous, and terrain with high relative altitude changes.
The Mars analog datasets also include high velocity, high travel distance, and higher altitude aspects.
Section~\ref{sec:groung_truth} discusses the methodology for the generation of ground~truth data.

Section~\ref{sec:providing_the_data} provides a brief outline of the data structure and~\ref{sec:usability_benchmark} shows two examples of processed data to prove the quality and usability of the data. More specifically, Section~\ref{sec:usability_transition} illustrates an exemplary multi-sensor fusion scenario that requires sensor switching for an outdoor to indoor transition scenario in the semi-urban area. Section~\ref{sec:usability_vio} shows a vision-only example, using VIO only with a comparison of the filter results globally aligned with ground~truth for a general comparison. 
Sections~\ref{sec:lessons_learned} and~\ref{sec:conclusion} complete the paper with lessons learned and the conclusion.
%

\section{Related Work}
\label{sec:related_work}
This section provides an overview of \ac{UAV} research datasets and how the presented work is positioned within this ecosystem.
The majority of open-source datasets in the robotics community focus on isolated research aspects
and most of them are tailored towards ground vehicles and aspects of autonomous driving.
This includes large-scale outdoor datasets such as KITTI \cite{Geiger2013} (1392x512~images~@\SI{10}{\hertz} and IMU~@\SI{10}{\hertz}) and the Oxford dataset \cite{Maddern2017_1year1000km} (1280x960~images~@\SI{16}{\hertz}, \ac{GNSS} and INS solution~@\SI{50}{\hertz} and no raw IMU data), with the later addition of sporadically sparse \ac{RTK} \ac{GNSS} for ground~truth \cite{Maddern2020}.
However, given the time at which the datasets were published, the provided data rates and image resolutions are lower compared to the sensor setup of the presented work.
Another ground vehicle dataset, targeted specifically for the SLAM and odometry community, is presented by \cite{Pire2019_therosariodataset}.
The dataset focuses on an agricultural environment and provides stereo imagery (672x376~@\SI{20}{\hertz}), IMU~(\SI{140}{\hertz}), and odometry data.
The dataset uses an \ac{RTK}~\ac{GNSS} for positional ground~truth but does not provide ground~truth for the global orientation.

Several indoor datasets are making use of local ground~truth in the form of \mbox{high-quality} SLAM in \mbox{post-processing}. This includes the TUM-LSI large-scale indoor dataset \cite{Walch2017} and \cite{Liu2022}, which adds a Leica station with fiducial markers at dedicated locations.
Another dataset, for a large-scale shopping mall scenery, introduced by Naverlabs \cite{Lee2021} makes use of \ac{SfM} in \mbox{post-processing} for the generation of ground~truth data.
The presented work does provide data to perform the same approach if desired. 
However, providing post-processed SLAM / \ac{SfM} data is not within the scope of this publication.

The next category of datasets concerns the indoor to outdoor transition aspect. Such datasets are mainly performed by handheld sensor suites.
\cite{Pfrommer2017_penncosyvio} introduced the PennCOSYVIO dataset with a sensor setup that extends a Google Tango platform with additional camera modalities (2x752x480~images~@\SI{20}{\hertz} and IMU~@\SI{200}{\hertz}).
The dataset features indoor and outdoor locations, but it does not make use of a \ac{GNSS} sensor. The ground~truth for this dataset is solely generated by utilizing pre-calibrated fiducial markers placed throughout the experiment area.
\cite{Schubert2018} later introduced the TUM dataset, also using a handheld sensor setup for indoor to outdoor trajectories without \ac{GNSS} information (1024x1024~images~@\SI{20}{\hertz} and IMU~@\SI{200}{\hertz}).
For this dataset, ground~truth information is only provided for the indoor segment using a motion capture system.
 
Finally, datasets for the development and proof of localization algorithms for UAVs also exist, but the focus does not lie on sensor degradation and sensor switching nor cross-domain operation.
The EuROC dataset introduced by \cite{Burri2016} focuses on various difficulty levels of VIO for indoor scenarios (2x768x480~images~@\SI{20}{\hertz} and IMU~@\SI{200}{\hertz}), with ground~truth generated by using a motion capture system and a Leica station.
The \mbox{UZH-FPV} drone racing dataset \cite{Delmerico2019} features dynamic flights for isolated indoor and outdoor trajectories (640x480~images~@\SI{30}{\hertz} and IMU~@\SI{500}{\hertz}/\SI{1000}{\hertz} respectively).
This dataset uses a motion capturing system for the ground~truth of the indoor trajectories, and SLAM for the ground~truth of the outdoor trajectories. 
Another interesting approach is the Blackbird dataset \cite{Antonini2020_ijrrblackbird}, which performs \mbox{real-world} flights and collects IMU~(\SI{100}{\hertz}) as well as motion capture information~(\SI{360}{\hertz}) to generate a multitude of \mbox{photo-realistic} vision streams, each for the same set of recorded trajectories.
UAV datasets for possible future real-world applications also exist; \cite{Rodrigues2021} generated a dataset for the analysis of energy consumption for a package delivery drone. However, the dataset does not provide image streams, and the accuracy of position ground~truth is only rated with $\pm \SI{2}{\meter}$~@\SI{10}{\hertz}.

Another dataset, mainly concerning the Mars analog contribution of the presented work, is the MADMAX dataset \cite{Meyer2021}. This dataset provides a sensor suite similar to the Mars-Rover but is performed in a handheld approach.
However, this dataset only provides lower rate IMU measurements~(\SI{100}{\hertz}) and lower rate image streams (1032x772~@\SI{15}{\hertz} and 2064x1544~@\SI{4}{\hertz}) compared to the presented INSANE dataset.
This is adequate, given that a rover platform does not perform agile trajectories.
The ground~truth for this dataset is generated by fusing measurements from two \ac{RTK} \ac{GNSS} units~(\SI{1}{\hertz}) with IMU measurements, resulting in a \SI{100}{\hertz} filtered ground~truth data stream.

\begin{figure}[!tb]
    \centering
    \includegraphics[width=1.0\linewidth, trim={0mm 0mm 0mm 0mm},clip]{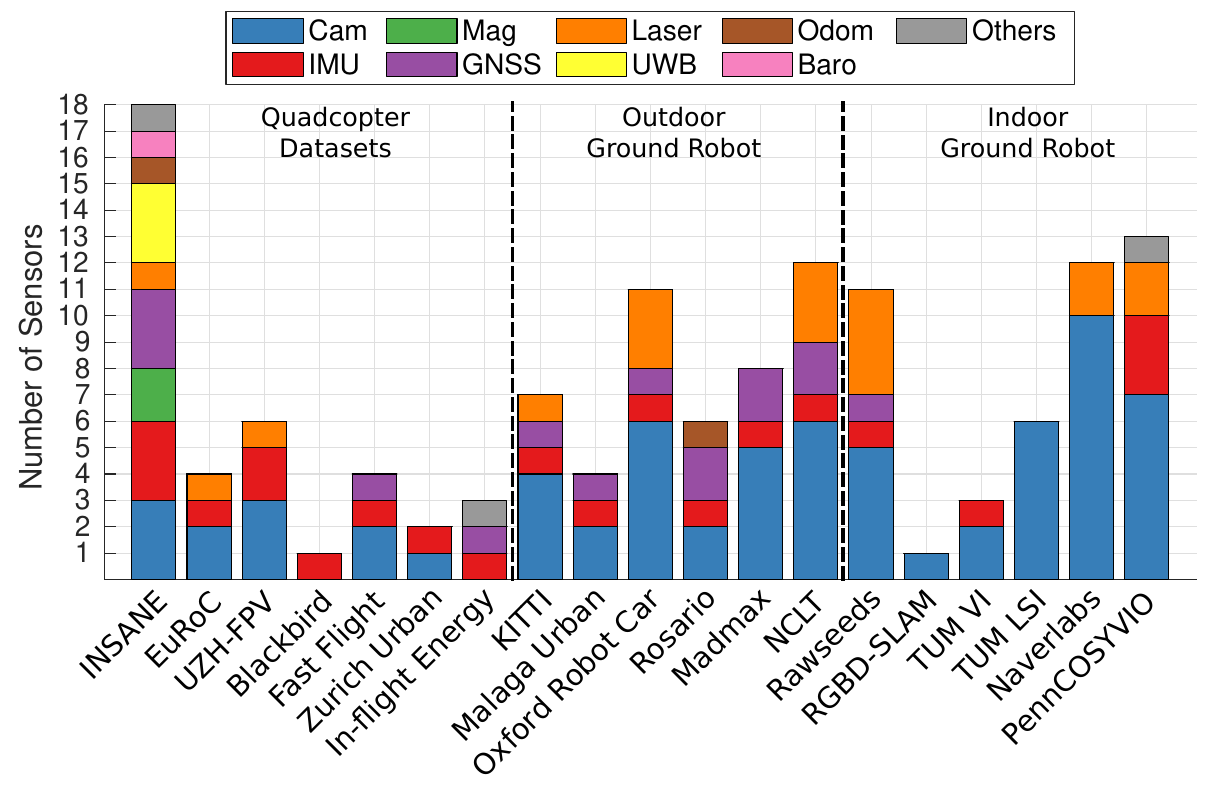}
    \caption{Overview of the number of sensors and their variety categorized by the type of robotic platform for state-of-the-art datasets. Corresponding tables, which compare the individual sensor modalities in more detail are provided in appendix \ref{sec:dataset_comparison}.
    }
    \label{fig:dataset_num_comp}
    \vspace{-5mm}
\end{figure}

Figure~\ref{fig:dataset_num_comp} shows a high-level overview of the number of sensors and the variety of sensor types across 19~datasets, comparing the INSANE and other datasets for quadcopters, outdoor ground-robots such as cars, and indoor ground and handheld robots. The tables in appendix \ref{sec:dataset_comparison} provide an additional and very detailed comparison between these datasets.

To the best of our knowledge, no UAV dataset concerning indoor-outdoor transitioning with aspects to real-world sensor degradation and an extensive sensor suite (see Tab.~\ref{tab:sensors}) with corresponding redundancies, various frame-rates and resolutions, and continuous global ground~truth for outdoor, indoor, and transition areas, such as presented by this work, exists at this point.

\begin{figure*}[htp]
    \centering
    \includegraphics[width=.8\linewidth]{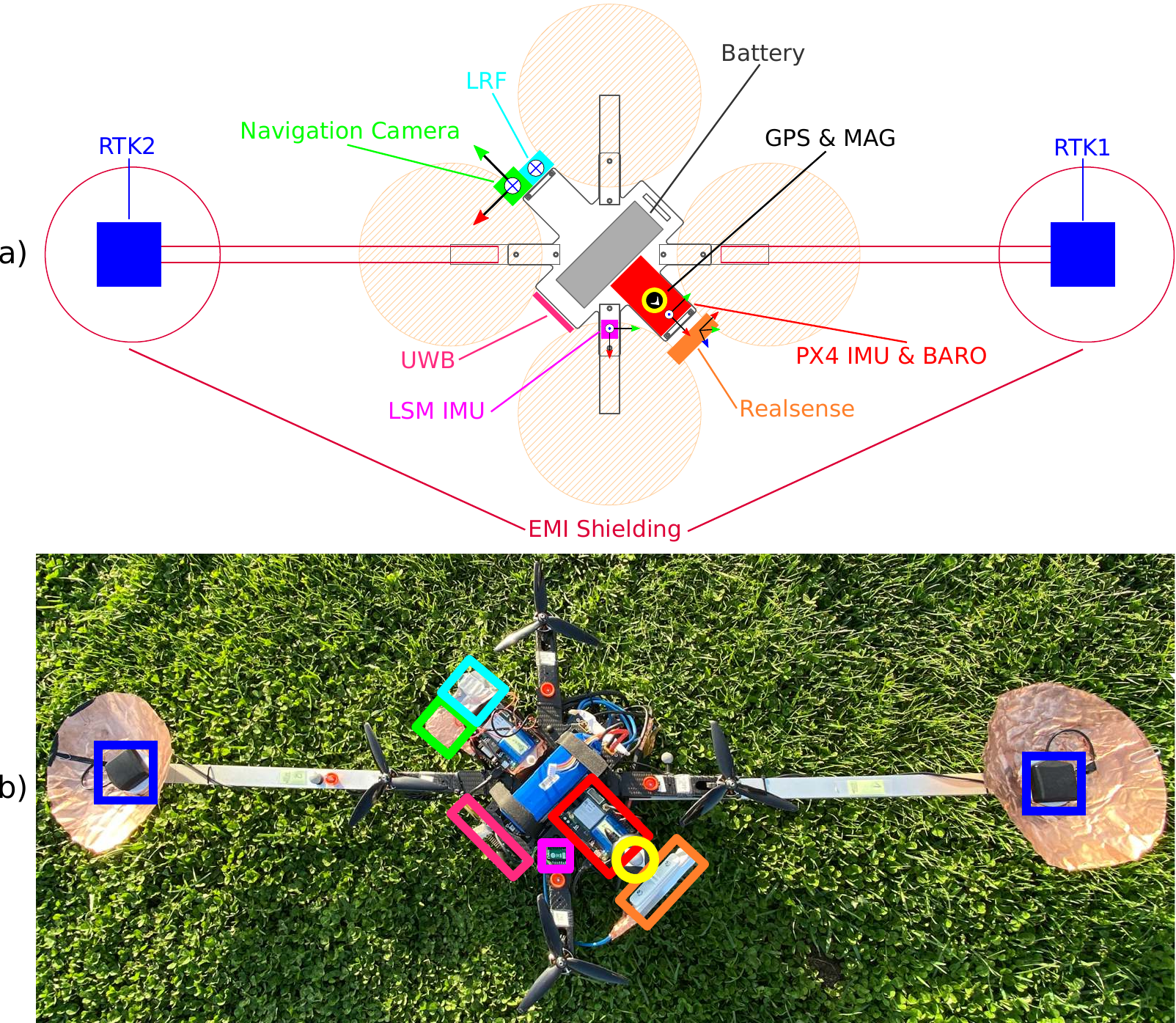}
    \caption{This illustration shows the vehicle design with coordinate frames for the placement and orientation of the sensors on the experiment platform in~\textbf{a)}, and the real vehicle with sensor position overlays in~\textbf{b)}.
    Accurate calibration of each sensor (extrinsics and intrinsics where applicable) is given for each scenario. The method on how each sensor is calibrated and the intent on the sensor placement is described in Section~\ref{sec:sensors}}
    \label{fig:experiment_platform}
    \vspace{-3mm}
\end{figure*}
%
\begin{table*}[htbp]
    \small
    \ra{0.8}
    \caption{The table describes the individual components of the sensor suite for the experiment platform. In particular, among others, it features three IMUs, three cameras and three GNSS sensors as well as two magnetometer.}%
    \vspace{-5mm}
    \label{tab:sensors}
    \begin{center}
        \begin{tabular}{ l c c p{6cm} }
        \toprule 
         \textbf{Sensor} & \textbf{Type} & \textbf{Rate [\si{\hertz}]} & \textbf{Description}\\
         \midrule
         \textbf{High Rate IMU} & LSM9DS1 & & \\
         IMU &  & 900 & Rigidly attached \\
         Magnetometer &  & 20 & \\
         \midrule
         \textbf{Pixhawk} \\
         \ac{IMU} & ICM20689 & 200 & Internaly dampened \\ 
         \ac{IMU} & BMI055 & 200 &  (Disabled) \\ 
         \ac{GNSS} &  & 5 &  \\ 
         Magnetometer & UST8310 & 80 &  \\ 
         Barometer & MS5611 & 20 &  \\ 
         Motorspeeds &  & 100 &  \\ 
         \midrule
         \textbf{RealSense~T256} \\
         \ac{IMU} & BMI055 & 200 & \\ 
         6DoF Odometry & V‑SLAM & 200 & \\ 
         \midrule
         \textbf{Cameras} \\
         Nav Camera & IDS UI-3270LE-M-GL  & 20 & CMOS Mono, 2056x1542, 3MP, global-shutter; Lens~BM4018S118C, FoV(D=\SI{126}{\degree}, H=\SI{101}{\degree}), aperture 1.8 \\ 
         Stereo Cam & RealSense~T256 & 30 & 848x800, global-shutter, \SI{64}{\milli\meter}, \SI{163}{\degree}~FoV \\ 
         \midrule
         \textbf{External Sensors} \\
         $2\times$ \ac{RTK} \ac{GNSS} & UBLOX C94-M8P & 8 & Coordinates and velocity \\ 
         \ac{LRF} & Garmin Lidar Lite v3 & 30 & \SI{40}{\meter} range, \SI{1}{\centi\meter} resolution \\ 
         $3\times$\ac{UWB} & Decawave TREK1000 & 7 & With additional vehicle marker \\ 
         $125\times$ Fiducial Marker & ArUCO &  & Rate is the same as Nav~Cam \\ 
         Motion Capture & Optitrack  & 300 & 37 camera dronehall setup \\ 
         Pulse Tachometer & Wachendorff PT99  & & RPM ground-truth \\
         \bottomrule
        \end{tabular}
    \end{center}
    \vspace{-2mm}
\end{table*}
%

\section{System Setup}
\label{sec:system_setup}

\subsection{Vehicle Configuration}
\label{sec:vehicle_config}
All datasets have been recorded with the flight platform shown by Figure~\ref{fig:experiment_platform}. The base of this platform is a commercially available carbon frame equipped with a minimal power unit, rotors, and a PixHawk4 autopilot.
The base frame was heavily extended and altered from its original. The final platform setup weighs \SI{3}{\kilo\gram} and carries a number of additional sensors (see Tab.~\ref{tab:sensors}).

\begin{figure}[htb]
    \centering
    \includegraphics[width=1.0\linewidth, trim=1cm 0.4cm 0.3cm 0.4cm, clip]{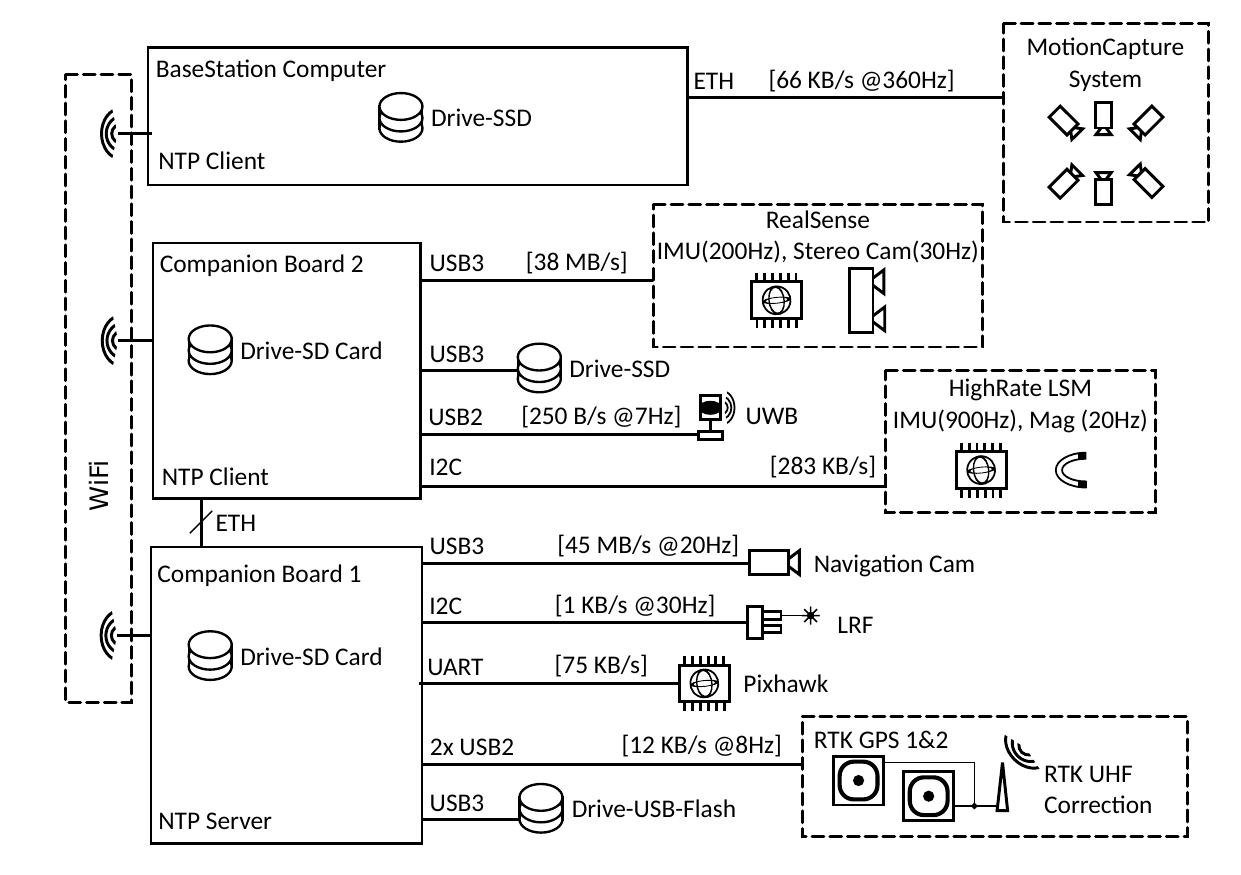}
    \caption{This diagram shows the embedded companion boards and sensor integration as well as storage and interface distribution according to individual sensor data rates and the available bandwidth of the interfaces from the embedded hardware.}
    \label{fig:system_diagram}
    \vspace{-8mm}
\end{figure}

The small size of the aerial vehicle constrains the amount of additional payload. However, the UAV setup needs to be able to process the sensor data during closed-loop experiments and needs to be able to record the data of the sensor suite in raw format without loss of information.
Because of this, the vehicle is equipped with two Raspberry~Pi4 companion boards. This allows for computational load balancing, interface bandwidth distribution associated with a specific sensor, and distributed sensor data storage. These three aspects of the vehicle system are shown as a block diagram in Figure~\ref{fig:system_diagram}.

It might be of interest to the reader that the vehicle was running an in-house developed and source-available\footnote{\url{https://github.com/aau-cns/flight_stack}} flight~stack \cite{Scheiber2022_cnsflightstack}, which is generalized and deployable across many standard computation platforms by utilizing a robust and versatile OS \cite{Stewart2021_skiffos}.

The vehicle includes additional \ac{EMI} shielding to allow optimal functioning of RF~sensitive components such as the \ac{GNSS} despite high-frequency data lines (further detailed in Section~\ref{sec:rtk_gnss}).
Additional dust shielding and individual cooling appliances were added to the flight platform for optimal operation in the hot and sandy environment posed during the Mars analog data recording sessions.

\subsection{Sensors}
\label{sec:sensors}
This section outlines the setup, calibration procedures, and individual aspects concerning the sensor suite. The sensors are summarized in Table~\ref{tab:sensors}.

\subsubsection{IMUs}
\label{sec:imu}
\begin{figure}[htb]
    \centering
    \includegraphics[width=1.0\linewidth]{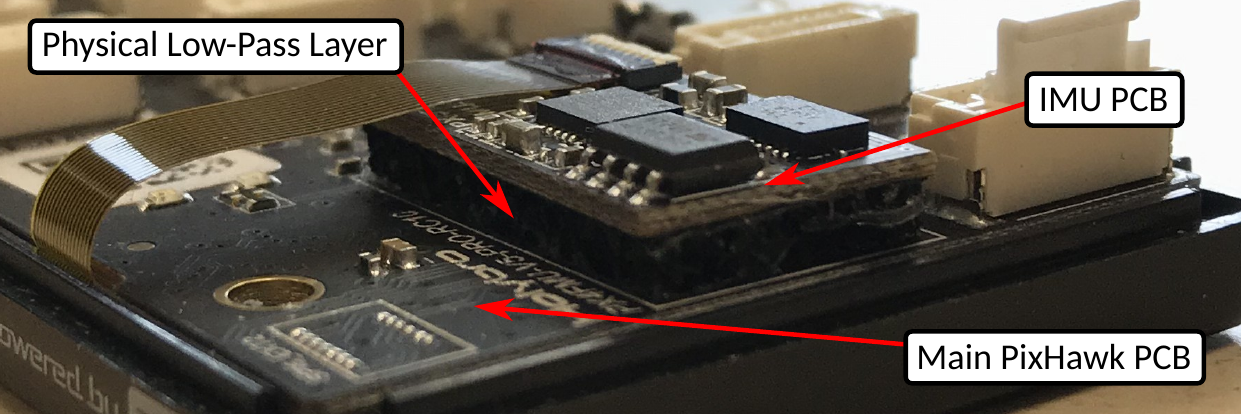}
    \caption{The sensor suite has three IMUs where each serves a different aspect. This figure shows the assembly of the primary IMU sensor within the PixHawk autopilot, on top of a damping layer that acts as a low-pass filter between the IMU and the main PCB board. This hardware filter is beneficial for closed-loop applications.
    However, in section~\ref{sec:vehicle_vibration_analysis}, we are carrying out a vibration analysis of the vehicle. This analysis made use of the rigidly attached high-rate IMU. A brief comparison of the dampened IMU and the \mbox{high-rate} IMU with applied filters, designed based on the vibration analysis, is described, and the interested reader is referred to~\cite{Steinbrener2022_imupropagation} for further use on the vibration data.}
    \label{fig:imu_damping}
\end{figure}

In general, the sensor suite includes four \ac{MEMS} IMUs.
Two \SI{200}{\hertz} IMUs are part of the Holybro~Pixhawk~v5 autopilot, namely the BMI055 from Bosch and ICM20689 from TDK.
One \SI{200}{\hertz} BMI055 IMU is part of the RealSense~T256 stereo camera, and one dedicated \mbox{high-rate} \SI{900}{\hertz} LSM9DS1~IMU from \mbox{ST-Electronics} is added by itself.
The two IMUs of the autopilot can not be distinguished and are actively switched based on a sensor voting scheme by default.
To enable a clear association between the IMU calibration and its bias, the BMI055~IMU was deactivated.
Another reason is that the RealSense~T256 already provides measurements of a BMI055 type IMU.
Thus, all active IMUs provided by the dataset are from individual manufacturers, providing a good range of different IMU characteristics.
In addition, it may be noted that the IMUs of the autopilot are hardware dampened, shown by Figure~\ref{fig:imu_damping} and further detailed in the vibration analysis Section~\ref{sec:vibration_test_bench}~and~\ref{sec:vehicle_vibration_analysis}, respectively.

The IMU sensor positioning is shown by Figure~\ref{fig:experiment_platform}.
The autopilot IMU, which is referred to as the main IMU, is positioned close to the vehicle's center, which should avoid amplified vibrations within the linear acceleration measurements. 

The RealSense camera is positioned forward-facing and tilted down for ideal stereo camera positioning.
Therefore, this IMU has the highest lever arm and receives the most amplification for resonances in terms of vibration.
Specific scenarios also showed undersampling effects. The measurement stream of this IMU can be seen as a challenging scenario for possible research.

The high-rate LSM9DS1 IMU is rigidly attached (not dampened) and provides measurements at more than~\SI{900}{\hertz} to support IMU filter applications or machine learning approaches, as shown by \cite{Steinbrener2022_imupropagation} for active vibration analysis and noise reduction.
%
The intrinsic calibration of the specific IMUs is done by performing the Allan variance method outlined by \cite{Freescale2015}.
IMU recordings with a length of five hours and the corresponding tools are open-sourced with the dataset.

\subsubsection{Magnetometer}
\label{sec:mag}
The helicopter platform hosts two independent magnetometers from different manufacturers for variation in sensor characteristics and redundancy. The first module is included in the PixHawk4 sensor suite (\SI{80}{\hertz}), and the second module is located within the external LSM9DS1 \ac{SoC} (\SI{20}{\hertz}).
For magnetometers, several aspects have to be considered: The intrinsic and extrinsic calibration of the sensor for its specific location on the experiment platform, the magnetic variation which depends on the geolocation, and local magnetic disturbances posed by the environment.
The dataset includes a default magnetometer calibration for this vehicle and dedicated magnetometer calibration datasets, which allows for different calibration methods by the user.

It is important to note that the magnetic variation at the locations in which the dataset was recorded are not negligible.
This variation depends on the geolocation and concerns the magnetic inclination (spherical~elevation) and the declination (spherical~azimuth), which build the local direction and strength of the magnetic field. The pertinent information is available  online\footnote{\url{https://www.ngdc.noaa.gov/geomag/calculators/magcalc.shtml}}.
Location-dependent variables and descriptions for the datasets can be found in Section~\ref{sec:environment_and_experiments}. Tables~\ref{tab:env_info_dronehall},~\ref{tab:env_info_airport} and~\ref{tab:env_info_israel} specifically address the magnetic variation.

\begin{figure}[t]
    \centering
    \includegraphics[width=1.0\linewidth, trim=3cm 0.8cm 3cm 1.2cm, clip]{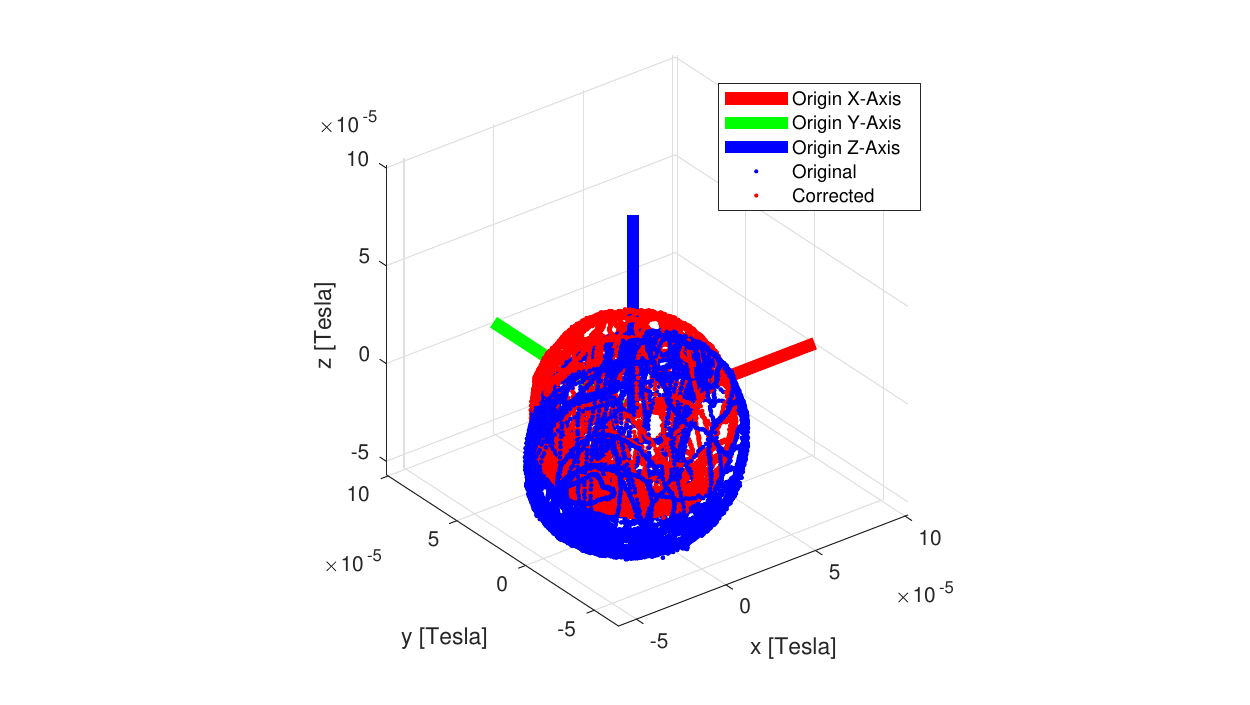}
    \caption{The image illustrates the offset for the magnetometer measurements from the PX4 platform, as an example on the magnetometer intrinsic calibration for hard- and soft-iron effects. The calibration is provided for all magnetometers.}
    \label{fig:mag_calibration}
\end{figure}

The provided data segments can be used to calculate the intrinsic and extrinsic calibration of a magnetometer with respect to an IMU sensor.
The intrinsic calibration of a magnetometer concerns hard-/ and soft-iron effects. Both effects are mainly dependent on the mounting location of the sensor.
Hard-iron effects cause a fixed offset in the magnetometer readings, while soft-iron effects affect the magnetic field strength and its direction locally.
As shown by Figure~\ref{fig:mag_calibration}, the red sphere shows raw magnetic measurements, and the blue sphere shows measurements that are compensated for their hard-/ and soft-iron effects.
The red sphere is not only shifted off-center due to hard-iron effects but also distorted in its spherical shape due to the mentioned soft-iron effects.

The calibration data was recorded in a non-occluded environment, and the vehicle was rotated around various rotational axes such that the magnetic vector, if projected onto a sphere, covers this sphere sufficiently.
At first, an ellipsoid fit for the raw data was performed and the center of mass for this ellipsoid was found.
This provides the general offset of the ellipsoid~$\mathbf{b}_\mathsf{ct} \in \mathbb{R}^{3\times3}$. Afterwards, a transformation was found which transforms the ellipsoid into a sphere~$\mathbf{t}_\mathsf{sp} \in \mathbb{R}^{3\times3}$.
Knowing both parameters, individual magnetometer measurements can be corrected by applying $\mathbf{m}_\mathsf{corr} = \mathbf{t}_\mathsf{sp} (\mathbf{m}_\mathsf{raw} - \mathbf{b}_\mathsf{ct})$.
Detailed information on this method can be found in \cite{Vasconcelos2011}.

After finding the intrinsic calibration, the extrinsic calibration between the IMU and magnetometer as well as the local magnetic inclination can be found. This is done by applying the method outlined in \cite{Papafotis2020}.
Individual data sequences were recorded which provide static rotations of the vehicle in all six rotational directions with respect to the gravity vector.
This data is processed with the cost function described in \cite{Papafotis2020} to find the most accurate rotation between the gravity and magnetic vectors, including the magnetic inclination.
The dataset is published with a default calibration for the vehicle setup and with a set of tools to recompute the calibration with potentially different parameters.

The data sequences for the transition experiments show an additional, real-world effect that we aimed to capture with our dataset. Because the indoor area and surrounding elements have metal structures, the magnetic field also changes depending on the location within the environment.
Figure~\ref{fig:mag_indoor_interference} shows how the magnetic field changes when entering the indoor area.
One possible approach on interacting with changes in the magnetic field is outlined in \cite{Brommer2020_mag}, which describes a method of detecting magnetic field changes with subsequent adaptation to the changing magnetic field without measurement rejection.

The magnetometer measurements for all flight sequences are not compensated nor calibrated.
This provides the possibility to include online self-calibration of magnetometer intrinsics, if desired.

\subsubsection{Cameras}
\label{sec:camera}
\begin{figure}[t]
    \centering
    \includegraphics[width=1.0\linewidth, trim=0cm 1cm 0cm 1cm, clip]{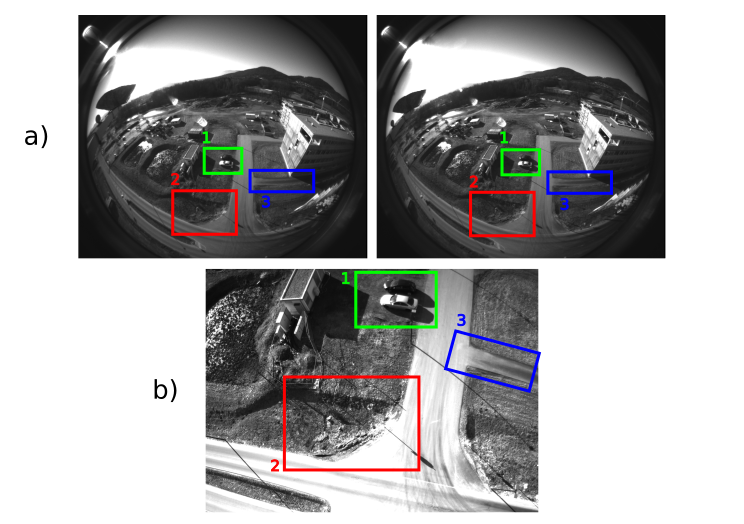}
    \caption{Example image of the outdoor dataset, showing the over-lapping FoV of the forward-down-facing stereo camera \textbf{(a)} and the downward-facing navigation camera \textbf{(b)}. Labels \mbox{1-3} show the same locations in the environment.
    \textbf{Note:} The image of the navigation camera is rotated by \SI{-90}{\degree} and cropped to the region of interest.}
    \label{fig:cam_overlap}
\end{figure}

The sensor suite includes a RealSense~T256 stereo camera (848x800~@\SI{30}{\hertz}) and an IDS~UI-3270LE-M-GL global-shutter 3MP navigation camera (2056x1542~@\SI{20}{\hertz}).
The navigation camera is facing downward and is aligned with an \acf{LRF} for associated pixel range information (see Sec.~\ref{sec:lrf}).
The stereo camera faces in flight direction and is tilted by \SI{\StereoCamAngle}{\degree}~towards the ground. The reason for this is threefold.
First, the stereo camera observes the horizon, and horizon detection can be used to improve attitude estimation.
Second, the forward-facing image stream, especially for the desert datasets, can be used to observe and map crater walls. A similar task might be included in future Mars explorations.
And third, starting at the height of \SI{2}{\meter}, the stereo camera and the navigation camera have an overlapping field of view, which allows for additional feature matching, as shown by Figure~\ref{fig:cam_overlap}.

Camera exposure time and gain were chosen such that motion blur is minimized. For the navigation camera, the exposure time was limited to~$<\SI{5}{\milli\second}$, and the gain was automatically adapted within a range that provides balanced brightness throughout an individual trajectory.
This especially concerns the transition datasets, which have three lighting conditions: Outdoor with natural light and radiant sun, the transition area with shadow casting and reflecting snow patches, and the indoor area with artificial light providing yet another light intensity.
The pure outdoor datasets, including the desert segments, make use of the same settings to reduce gain where flat patches of reflecting areas might raise the brightness of the image over a suitable threshold.
The camera settings for the stereo camera are left to default values, which provide suitable imagery throughout all scenarios. 

The timestamps of the navigation camera are provided by an internal clock of the camera, which is synchronized to the host system on startup. Thus, the timestamps reflect the time at which the image was taken by the camera module.

The dataset further provides dedicated recordings for the calibration of both cameras. This includes measurement streams of all IMUs and an image stream for the dedicated camera observing a checker/fiducial marker board.
This data can be used to generate the intrinsic and extrinsic calibration of the cameras with respect to any IMU by using the Kalibr tool \cite{Rehder2016_extendingkalibr}.

\subsubsection{Laser Range Finder}
\label{sec:lrf}
All datasets feature a \SI{30}{\hertz} data stream of a Gamin Lidar~Lite~v3, which is a laser-based range measurement sensor. This sensor provides distance to ground measurements at a resolution of \SI{1}{\centi\meter} with \SI{2.5}{\centi\meter} standard deviation and operates at a distance of up to \SI{40}{\meter}, according to the manufacturer's description. It was noticed throughout the recording of the datasets that this sensor provides sporadic zero measurements beginning at distances of 25-30~meters depending on the properties of the ground surface. The detection and rejection of these measurements is thus straightforward.

\begin{figure}[t]
    \centering
    \includegraphics[width=1.0\linewidth]{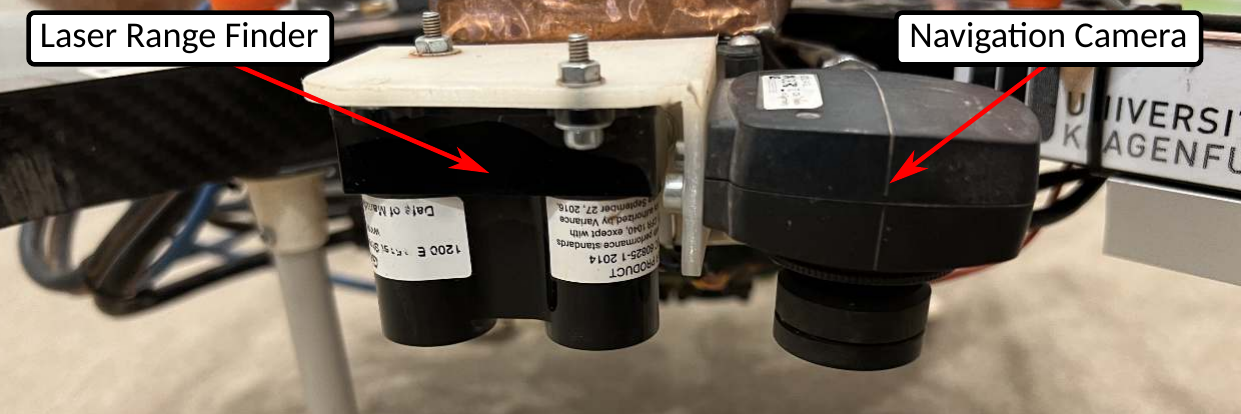}
    \caption{This illustration shows the co-mounted camera and laser range finder setup, which provides the same sensor setup as for the Mars-Helicopter ingenuity described by \cite{Grip2019}.}
    \label{fig:cam_lrf}
\end{figure}

This sensor is co-mounted with the high-resolution navigation camera to provide the same setup which is used by the Mars-Helicopter Ingenuity \cite{Bayard2019}.
The \ac{LRF} is mounted facing down in the same direction as the navigation camera (see Fig.~\ref{fig:cam_lrf}) to provide distance-to-ground measurements for a known pixel cluster within the navigation image.

The dataset provides two options for the calibration of the rotation between the camera and the \ac{LRF}.
The calibration routine for the first option provides an image stream with a visible checker/fiducial target board for pose tracking and associated \ac{LRF} and IMU measurements. The dataset contains measurements while the system was excited in all dimensions. A calibration can be performed with the Kalibr tool \cite{Rehder2016_extendingkalibr}.

The second option is to use the data for the static setup in which the camera and the \ac{LRF} are facing a wall at a defined distance. With the help of an indicator card, the spot in which the infrared \ac{LRF} laser hits the wall is made visible and can thus be marked in the image and associated with a pixel location.
A calibration that was generated with the second method is provided with the dataset.

\subsubsection{Real-Time Kinematic (RTK) GNSS}
\label{sec:rtk_gnss}
The sensor suite includes two \mbox{off-the-shelf} UBLOX \mbox{C94-M8P} \ac{RTK} \ac{GNSS} development boards with standard ceramic patch antennas.
An \ac{RTK} \ac{GNSS} system requires a static base station that communicates corrections to the \ac{GNSS} modules mounted on the vehicle. This results in highly accurate \ac{GNSS} measurements with an accuracy of \SI{1}{\centi\meter} to \SI{16}{\centi\meter} standard deviation.
For the available datasets, the threshold for the calibration of the \ac{RTK} base station was set to \SI{0.5}{\meter}.

The UBLOX \mbox{C94-M8P} module provides a \ac{UHF} module for the communication between the base station and the receivers on the vehicle. This communication channel was replaced by ZigBee modules for an extended range and reliability of the presented setup.

An \ac{RTK} \ac{GNSS} module has different modes of operation, which indicate the level of accuracy in position.
The three primary modes are non-RTK standard \ac{GNSS}, RTK float, and RTK fixed mode.
Non-RTK represents the standard \ac{GNSS} information without additional corrections.
The RTK-float state signalizes that no unique solution for the current constellation exists.
This state still benefits from \ac{RTK} corrections and provides measurements with \SI{10}{\centi\meter} to \SI{16}{\centi\meter} accuracy.
The RTK-fix is the most accurate state and provides an accuracy of below \SI{1}{\centi\meter}.
The receiver's status depends on a multitude of factors, such as the number of satellites, the time each satellite was tracked, and the SNR posed by the environment.

The majority of the recorded datasets provide measurements with RTK-fix measurements. However, certain conditions can cause sporadic RTK-float states.
The \ac{GNSS} receivers also provide an estimate for the measurement covariance for significantly more detail on the measurement accuracy.
This information is also provided with each dataset. We used the binary information of fixed and float \ac{RTK} states for the detection of \ac{EMI} issues.

\begin{figure}[h]
    \centering
    \includegraphics[width=1.0\linewidth, trim=0.8cm 1.2cm 0.8cm 1.2cm, clip]{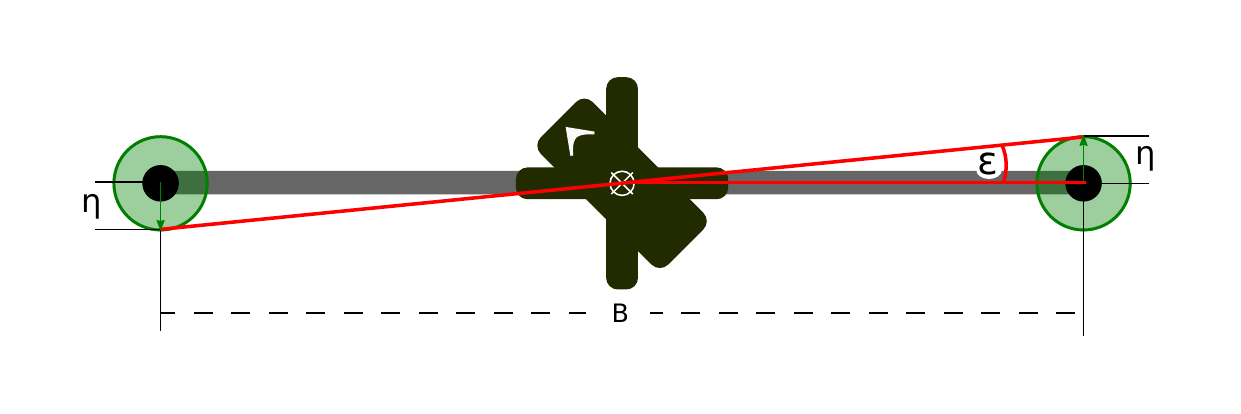}
    \caption{Rotational dual \ac{GNSS} error $\epsilon$ based on measurement accuracy $\eta$ and the antenna baseline $B$.}
    \label{fig:gnss_rot_error}
\end{figure}

As illustrated by Figure~\ref{fig:experiment_platform}, the two \ac{GNSS} antennas are placed on aluminum rods with a distance of \SI{1.2}{\meter}, centered at the midpoint of the vehicle and rotated by \SI{45}{\degree} with respect to the front of the vehicle represented by the positive x-axis of the main IMU.
The reason for the large baseline between the \ac{GNSS} antennas is that the \ac{GNSS} measurements are used for the position and rotational ground~truth. Given the setup shown by Figure~\ref{fig:gnss_rot_error}, an accuracy of $\eta = \SI{1}{\centi\meter}$ for one antenna, and the baseline of $B = \SI{1.2}{\meter}$, the rotational error in the worst case is $\epsilon = \SI{0.95}{\degree}$.
The method used for ground~truth generation is detailed in Section~\ref{sec:groung_truth}.

We want to briefly address the topic of \ac{EMI} because the vehicle carries numerous components such as processors, RF communication, and DC/DC power converters which pose the potential for \ac{EMI} with the off-the-shelf GNSS antennas:
\ac{EMI} in the context of high-frequency communication channels such as Bluetooth and especially USB3 has been discussed and analyzed in the literature \cite{Lin2014}.
Because the \ac{GNSS} ceramic patch antennas for this system are standard consumer-grade products, the antennas do not make use of additional active interference rejection and provide no dedicated shielding against \ac{EMI}.
Initial \ac{GNSS} quality tests showed impaired signal quality mainly due to the low SNR of individual satellite signals.
This is caused by the onboard electronics, such as the non-shielded computation boards, sensors, and high-frequency data lines hosted by the platform.
Thus, the antennas required significant additional shielding (see Fig.~\ref{fig:experiment_platform}) to reach an SNR, which allows the operation with high accuracy.

\begin{figure}[t]
    \centering    
    \includegraphics[width=1.0\linewidth, trim=0.2cm 0.4cm 0.2cm 0.5cm, clip]{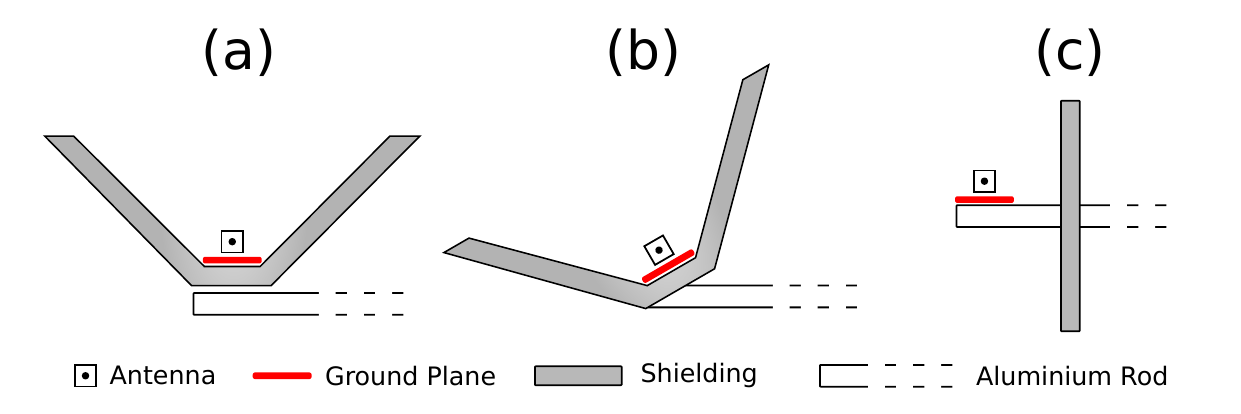}
    \caption{Tested \ac{EMI} preventive variations for dedicated \ac{RTK} Antenna shielding to archive sub-centimeter position accuracy and subsequent orientation accuracy for the ground~truth method described in Section~\ref{sec:ground_truth_vehicle_pose}. \textbf{a)}~upward-opening with $45\unit{\degree}$~side and base shielding, \textbf{b)}~similar to \textbf{(a)} but antenna and shielding tilted \SI{40}{\degree}~in the vertical plane, \textbf{c)}~fully vertical shielding, no ground encapsulation besides antenna ground plane.}
    \label{fig:rtk_shielding}
\end{figure}

Different shielding designs, made from copper sheets and constructed by hand, as shown by Figure~\ref{fig:rtk_shielding} have been tested.
In all tests, shielding variation a) showed the most improvement for the SNR. Option c) mainly failed due to possible ground reflections of vehicle-centered \ac{EMI} sources.
Option b) showed good static results, but tests showed that long-term tracking of satellite signals was interrupted during rotational movement.
The \ac{RTK} receiver requires observations of satellite signals over more extended periods of time. Shielding option b), however, causes a barrier of signals to one side of the hemisphere, which changes based on the rotation of the vehicle.
Thus, shielding variation a) was ultimately used for all data recordings.
In addition to the apparent antenna shielding, the final setup required individual shielding of USB3 connection points and additional USB3 inline low-pass filter\footnote{Wuert Elektronic 829993STICK} as well as shielding and partial enclosure of computational platforms and the \ac{GNSS} receivers.

As mentioned in the previous section, the quality of the \ac{GNSS} also depends on the environment.
The transition data was recorded in a semi-urban environment prone to multi-path effects and a higher noise floor.
On the contrary, the Mars analog datasets recorded in the Negev desert have a very low noise floor.
Such circumstances can be observed in the corresponding covariances of the \ac{GNSS} measurements.
However, we made sure that the data in the datasets always have high-quality \ac{GNSS} measurements and thus highly accurate ground~truth.

\subsubsection{Fiducial Marker}
\label{sec:fiducial_marker}
The transition and indoor datasets provide fiducial marker landmarks. The primary purpose of these markers is to provide ground~truth pose information in the transition area.
When moving from outdoors to indoors, the transition area poses the challenge in which highly accurate \ac{GNSS} position measurements degrade in terms of accuracy as the vehicle moves closer towards the building.
At the same time, the motion capture system does not detect the vehicle yet.
Thus, a field of 108~fiducial markers on eight rigid platforms is used to generate ground~truth information for this trajectory segment. Each marker is a uniquely identifiable ArUco marker\footnote{ArUco Dict. 7x7 250 } generated with the library introduced in \cite{Garrido-Jurado2016}.
The markers become visible while the \ac{GNSS} accuracy is still sufficient. During a handover phase, the \ac{GNSS} and marker pose information is aligned to provide a common ground~truth reference frame.
After this handover phase, only the marker poses are used to generate ground~truth information until the vehicle reaches the inside of the building and the area that is covered by the motion capture system.
This represents the second handover phase in which a transition from marker ground~truth to the motion-capturing system is performed.

The system uses two board types, five narrow-field and three wide-field marker boards. The difference between these boards is the marker size.
The wide field boards carry a $\SI{50}{\centi\meter}$ marker which becomes visible at the height of~\mbox{$<\SI{10}{\meter}$}. These boards are used for the outdoor area.
The narrow field boards feature smaller markers to cover lower-altitude detections during the indoor transition phase.

The fiducial marker field is calibrated based on a dedicated recording with the navigation camera, which covers all markers across boards.
Marker poses $\mathbf{T}_{\text{cam}\_\text{marker}}$ are expressed with respect to the navigation camera frame. First, all transformations between visible markers in each image $\mathbf{T}_{ij} = \mathbf{T}_{\text{cam}\_i}^{-1}~\mathbf{T}_{\text{cam}\_j} \text{~with~}j<i~\in\mathbb{N}$ are extracted. This condition is applied to reduce complexity.
Once all images have been analyzed, the individual transformations $\mathbf{T}_{ij}$ are filtered for outliers not laying within $1\sigma$ standard deviation, and the mean is generated. 

A main marker is defined, which builds the reference frame for all other markers.
All individual calibrations, which may not be observed together with the main marker, are built by forming a graph search with the shortest path regarding the number of needed translations between the main marker and individual markers.
Since this can include erroneous translations, the pose of each marker with respect to the main marker is improved by building multiple randomized paths between the main and individual markers, which are then averaged using the geometric median.
Finally, each marker is referenced to the main marker. The board which is carrying the main marker is associated to a motion-capturing object. Thus, all markers can be expressed with respect to the motion capture system.
This static calibration (visualized by Fig.~\ref{fig:fiducial_marker_calibrated_field}) is used to generate poses for the vehicle in the transition scenario case.

\begin{figure}[!b]
    \centering
    \includegraphics[width=1.0\linewidth, trim=1.6cm 1cm 1.4cm 0.2cm, clip]{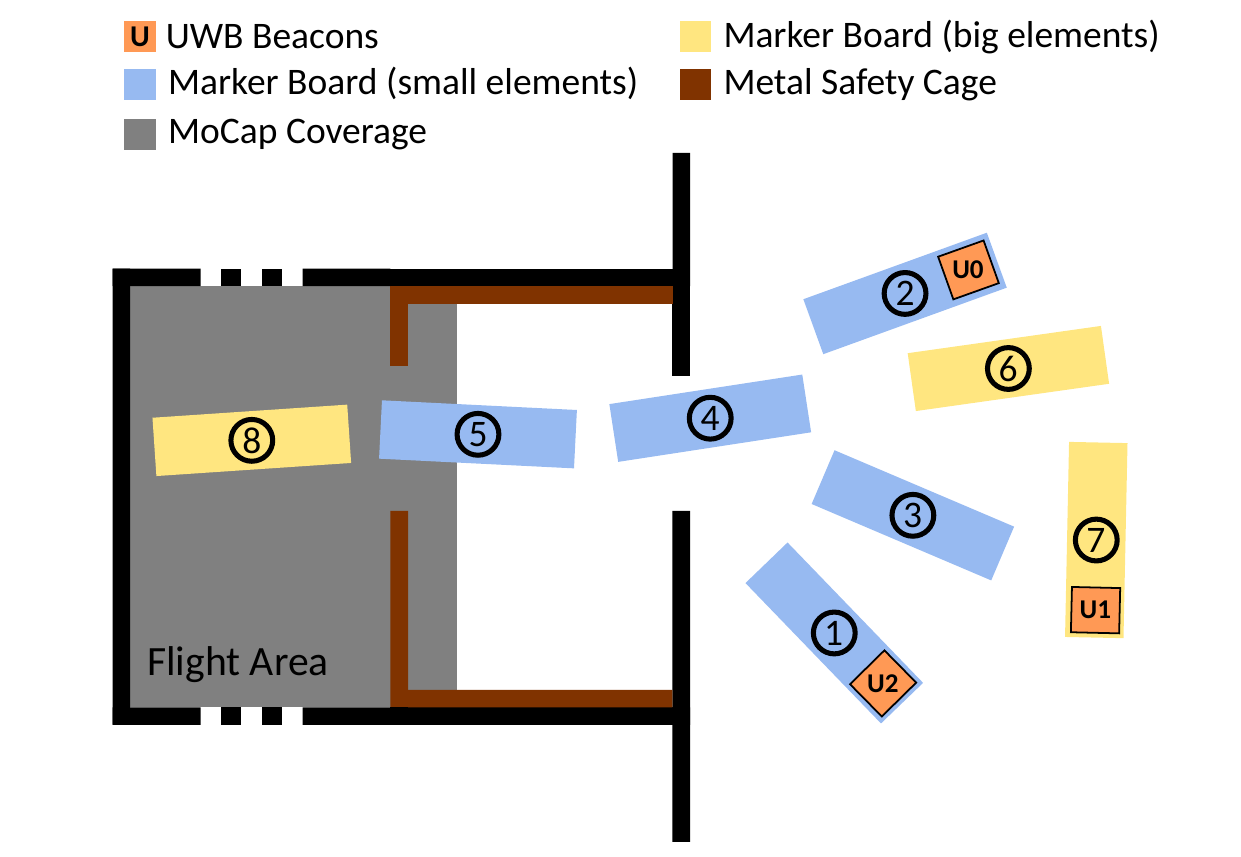}
    \caption{ArUco fiducial marker field design for the outdoor-indoor transition area at the Dronehall Klagenfurt. Small markers on the edge of each board provide higher accuracy for board-to-board calibration. Indoor markers are small to medium-sized due to the close proximity of the vehicle during the experiment. 
    The outdoor marker boards carry markers that allow reliable detection up to \SI{10}{\meter} altitude. This Setup ensures sufficient overlay in which high \ac{RTK} \ac{GNSS} quality is given, and the navigation camera can detect the marker. The image further shows the motion capture coverage and the placement of the \ac{UWB} modules.}
    \label{fig:fiducial_marker_distribution}
\end{figure}

The dedicated layout for this location is shown by Figure~\ref{fig:fiducial_marker_distribution} and the location of the markers after calibration is shown by Figure~\ref{fig:fiducial_marker_calibrated_field}.
A second purpose for these boards is the reference of \ac{UWB} modules located outside the building and mounted to the marker boards. The \ac{UWB} modules are placed on three boards and known locations for \ac{UWB} anchor calibration.

Markers can also be used as a visual aid and are present for the indoor recordings for this purpose as well.
Prepared camera relative poses are provided with the datasets as well as the marker calibrations.
All tools to export this data from raw data streams are open-sourced with the dataset.

\begin{figure}[t]
    \centering
    \includegraphics[width=1.0\linewidth, trim=1.8cm 0cm 1.8cm 0.8cm, clip]{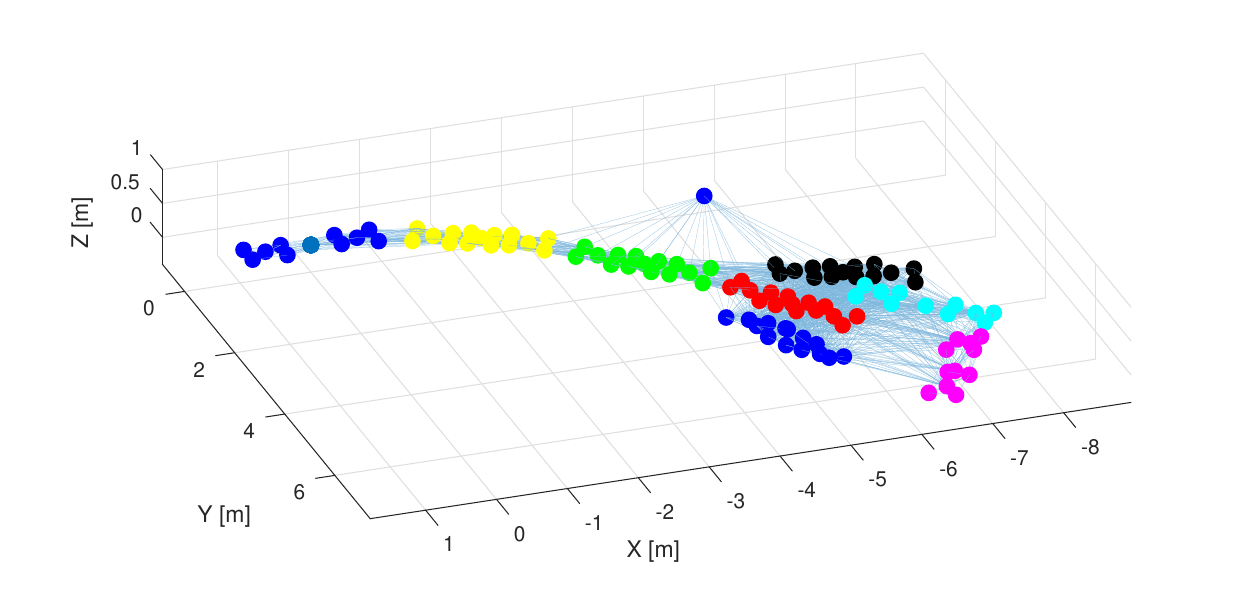}
    \caption{The graph shows the placement of individual fiducial markers and their respective carrier boards in the outdoor/indoor transition area after the calibration routine. Colored dots represent individual markers detected, a dedicated color is used for each carrier board, and the lines connecting the dots represent observed relative translations between markers during the calibration procedure. The designed layout is shown in Figure~\ref{fig:fiducial_marker_distribution}.}
    \label{fig:fiducial_marker_calibrated_field}
\end{figure}

\subsubsection{Ultra-Wideband Modules}
\begin{figure}[b]
    \centering
    \includegraphics[width=1.0\linewidth]{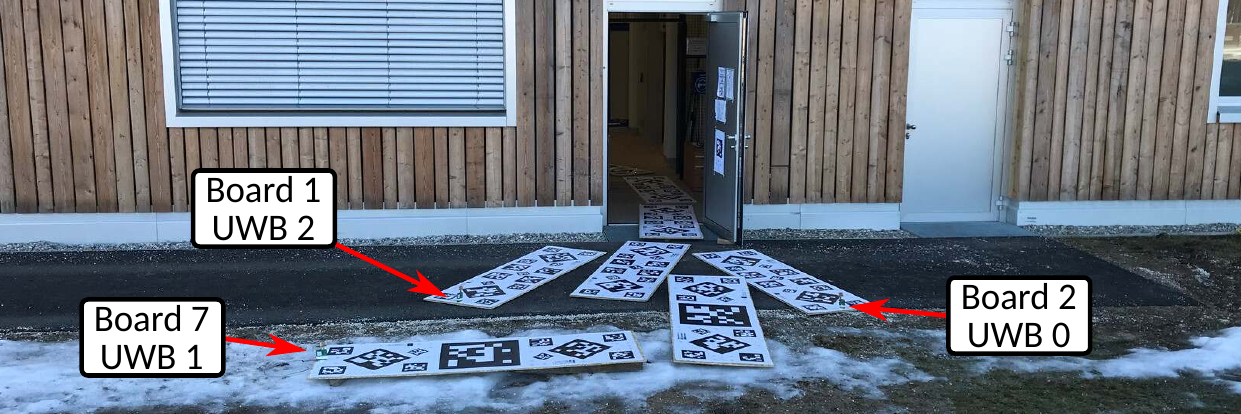}
    \caption{The entrance of the Dronehall, which the aerial vehicle is flying through for the recording of the transition phase datasets. The image illustrates the setup of the marker boards and the distribution of \ac{UWB} modules.}
    \label{fig:dronehall_entrance}
\end{figure}

The indoor and transition datasets also feature \ac{UWB} measurements. In the case of the transition dataset, the \ac{UWB} modules are placed close to the building entrance and can be used for position estimates in the area in which the \ac{GNSS} signals are degrading.
For this purpose, the vehicle is carrying a main anchor, and three additional \ac{UWB} anchors are placed on the ground. The \ac{UWB} positions are shown in Figure~\ref{fig:fiducial_marker_distribution} and \ref{fig:dronehall_entrance}. Each \ac{UWB} measurement is associated with a tag.

The position of the ground \ac{UWB} anchors is determined by known positions on the fiducial marker boards, which can be related to the indoor motion capturing system (see Sec.~\ref{sec:fiducial_marker}).
The range in which the \ac{UWB} modules provide measurements depends on environmental occlusions. While the vehicle is in the air, the first measurements are received from a distance of~\SI{30}{\meter}. A reliable stream of measurements is available at a range of~$<\SI{15}{\meter}$ between \ac{UWB} anchors. This includes partial areas of the indoor area.

\subsubsection{Computation Module Time Synchronization}
\label{sec:module_synchronization}
The system diagram in Figure~\ref{fig:system_diagram} shows that the vehicle uses two onboard computational modules for the acquisition of sensor data. These two modules are used for all experiments. A third computational instance is dedicated for the recording of motion capture data and is present for all experiments that involve the indoor area.
Because multiple modules are used, inter-module time synchronization is essential to ensure coherent data association of isolated modules.

All platforms run an NTP~server through Chrony\footnote{https://chrony.tuxfamily.org/}, which is set up to synchronize the system clocks at a higher rate compared to default settings.
The NTP time server/reference is set up to be on module~1 of the vehicle companion boards. Module~2, and the motion capture system run an NTP~client, which is continuously synchronizing to the reference time source.

\begin{figure}[htb]
    \centering
    \includegraphics[width=1.0\linewidth]{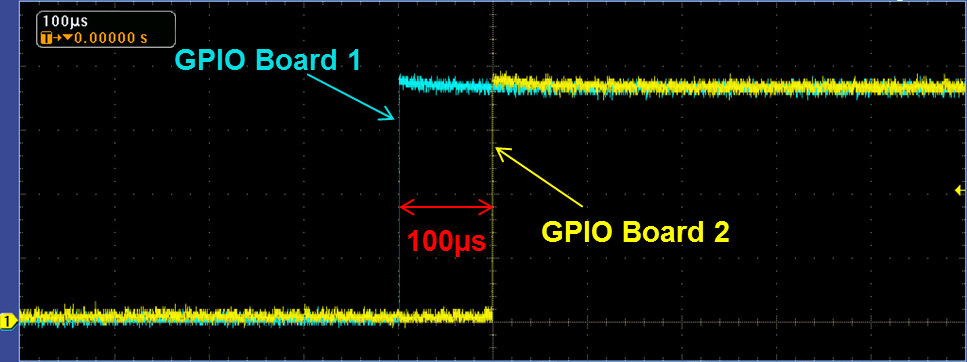}
    \caption{Validation of the time synchronization for the onboard computation modules.}
    \label{fig:gpio_time_sync}
\end{figure}

We performed a dedicated test for the validation and evaluation of the quality for the time synchronization.
This test was performed by utilizing the \acp{GPIO} of the two onboard modules, which were setup to generate a digital signal with rising edges at the same time, based on the system clock. The signal of both modules was observed with an oscilloscope.
This experiment setup showed that the synchronization of the system time is accurate to \SI{100}{\micro\second} (see Fig.~\ref{fig:gpio_time_sync}) using an NTP server with continuous synchronization steps.

In addition to the synchronization of the computation modules, we also want to address the synchronization of the IMUs and the RealSense T256 module.
%
The synchronization of the PX4 IMU is done software-based. The PX4 autopilot continuously synchronizes its internal clock with the computation module (host) using an NTP-like protocol for the correction of its internal clock. Timestamps of all PX4 sensor readings are then done onboard the PX4 and sent to the host with the corresponding system time.

The LSM9DS1 IMU is a single stand-alone chip which is readout via I$^2$C. This device does not have an internal clock that can be synchronized. Thus, the timestamp generation of the LSM9DS1 is done as follows. The driver requests a measurement which triggers the acquisition of the IMU readings. The driver is immediately triggered as soon as an IMU reading was received on the I$^2$C bus.
The time difference between the request and the response is called the round trip time and is used to compensate for the communication delay, which allows for a more accurate timestamp 
$t_\mathsf{stamp} = t_\mathsf{request} + 0.5~(t_\mathsf{response} - t_\mathsf{request})$.

The synchronization of the RealSense T256, according to its documentation\footnote{\url{https://github.com/IntelRealSense/librealsense/blob/v2.53.1/doc/t265.md}}, is done as follows.
The T256 uses a continuous synchronization method, which provides an accuracy of \SI{1}{\milli\second} between the internal timestamps of the T256 and the host.
Timestamps of the IMU data and the camera images are based on the common internal clock of the T256 and provided as the hosts system clock time.

Overall, please note that the individual sensors are not hardware synchronized. A software synchronization with an accuracy of \SI{100}{\micro\second} provides a sufficient base for sensor synchronization. We want to point out that hardware synchronization is generally not standard practice and not always possible. Novel approaches should thus not rely on synchronously triggered sensor information and allow for online estimation of delays.

However, accurate ground~truth for the state of a vehicle is important for a dataset, and we perform additional synchronization for the sensors that are used to generate ground~truth information in post-processing (see Sec. \ref{sec:gt_time_sync}).
The post-processing tools for the synchronization of the ground~truth sensors are publicly available and can be used for the synchronization of other elements.

\subsection{Vibration Test Bench}
\label{sec:vibration_test_bench}
\begin{figure}[htb]
    \centering
    \includegraphics[width=1.0\linewidth]{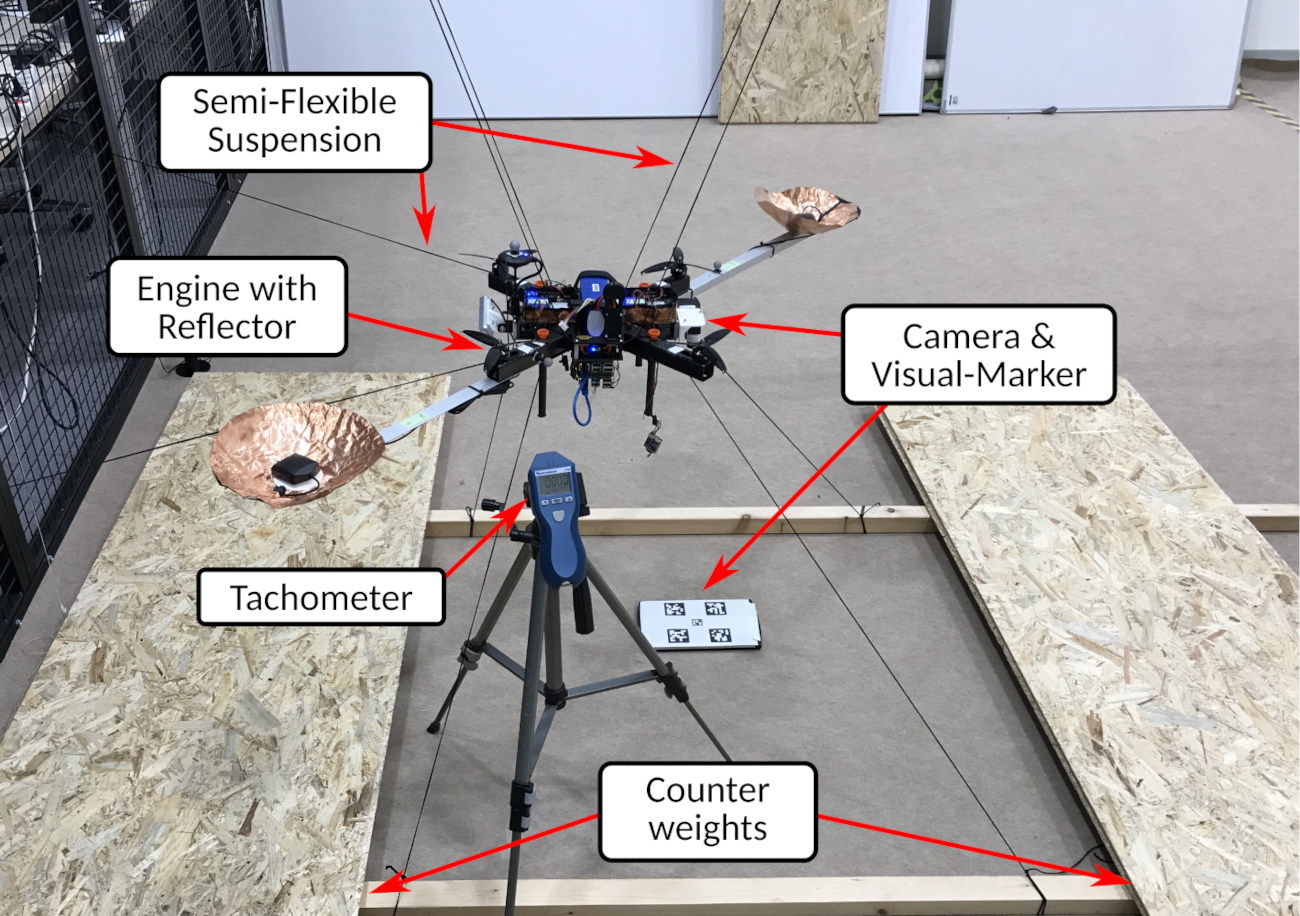}
    \caption{The image shows the vibration test bench to which the vehicle is semi-rigidly attached. Variations of static RPM and RPM sweeps of the engines provide a baseline for the vibration characteristics of the vehicle for subsequent analysis of the three IMU signals in nominal condition.}
    \label{fig:vibration_test_bench}
\end{figure}

\begin{figure}[htb]
    \centering    
    \includegraphics[width=1.0\linewidth, trim=1.5cm 0.8cm 1.3cm 0.9cm, clip]{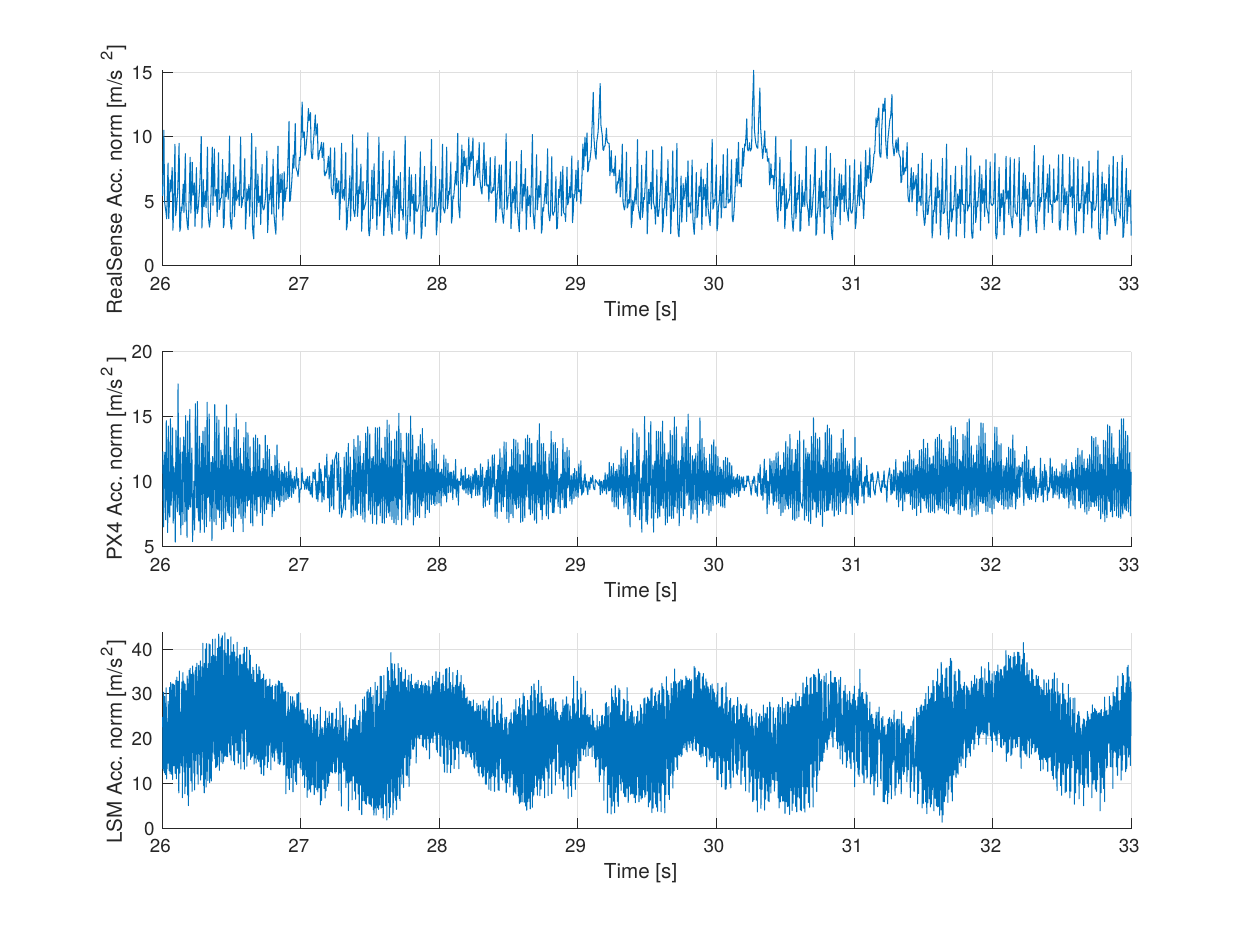}
    \caption{This plot shows a comparison for the norm of the linear acceleration signal from the BMI055 RealSense IMU \textbf{(top)}, the ICM20689 PX4 IMU \textbf{(middle)}, and the LSM9DS1 IMU \textbf{(bottom)}.
    The vehicle was suspended in the test bench shown by Figure~\ref{fig:vibration_test_bench}, and the motor speeds were set to $15500$~RPM.
    The RealSense IMU was mounted off-center and thus with a lever arm, and the PX4 IMU was mounted in the center of the vehicle. Both IMUs have a measurement rate of \SI{200}{\hertz} and show under-sampling effects with corresponding correlations.
    The LSM9DS1 IMU was also mounted off-center but has a measurement rate of \SI{900}{\hertz}, which results in a more accurate sampling of the vibration signals.}
    \label{fig:undersampling_realsense_imu}
\end{figure}

The dataset also provides additional data for vehicle-specific vibration analysis. This data is especially interesting for advanced IMU pre-filtering.
The high-frequent vibration characteristics of the vehicle showed under-sampling effects for the low-rate IMUs (RealSense and PX4~@\SI{200}{\hertz}). The RealSense is also mounted off-center, which increases this effect. Further investigation also showed that the IMU of the PX4~autopilot was hardware dampened (see Fig.~\ref{fig:imu_damping}).
To capture all vibration dynamics of the vehicle without hardware dampening and at a high enough sampling rate, a dedicated rigidly attached high-rate \SI{900}{\hertz} IMU was added to the sensor suite.
Figure~\ref{fig:undersampling_realsense_imu} shows a signal comparison between all IMUs. Please note that the Figure shows the norm of the linear acceleration per IMU, the change in magnitude for the LSM9DS1~IMU can be explained by the lever arm. The change in magnitude for the RealSense IMU can be caused by the lever arm as well as hardware dampening and settings from the manufacturer, which could not be changed to RAW.

The resulting dataset is suitable for an in-depth analysis of conditional sensor signal characteristics, and enables the development of adaptive noise canceling and other IMU pre-filtering techniques for improved trajectory tracking.

Data for the vibration analysis was recorded with the test bench shown in Figure~\ref{fig:vibration_test_bench}.
The flight platform was restrained by ten ropes, eight ropes to restrain the vehicle in the vertical axis, and two ropes to further restrain rotations in the horizontal plane.
The goal was to provide a semi-rigid constraint that prevents large movement of the platform but allows the nominal vibration characteristics of the vehicle that it exhibits during unconstrained, real-world flight.

The recorded data for this scenario includes the three IMU data streams, magnetometer, pressure, downward-facing \ac{LRF}, and nominal motor speeds.
The PX4 firmware was adapted to stream the nominal RPM values reported by the internal mixer module.
A tachometer, with provides the direct RPM of a single-engine, is used to scale the nominal motor speeds reported by the autopilot.
In addition, the video streams of both cameras are recorded as well. The downward-facing navigation camera records a small five-element fiducial marker board, and the forward-tilted stereo camera observes another fiducial marker board.
The pose information of the marker throughout the vibration data sequences can be used to refactor the vehicle pose data and exclude low-frequent movement - if desired.
It further serves the purpose of verifying the credibility of the camera sensor mounting design.

The vibration data is recorded in nine segments with motor speeds ranging from \SI{10}{\percent} to \SI{100}{\percent} of the full nominal motor speeds.
The vehicle is static before each sequence, and the motors are at a full stop before and after each segment is recorded.
All four engines receive the same thrust signal through software commands, which ensures fixed and equal RPMs for all motors during the recorded sequence.
A more detailed analysis of the vibration data is presented in Section~\ref{sec:vehicle_vibration_analysis}.

\section{Environments and Experiment Data}
\label{sec:environment_and_experiments}
The dataset provides experiment data for three different environments. This includes the environmental transition scenario at the campus at the University of Klagenfurt, a solely outdoor setup at a model airfield, and an outdoor Mars analog setup at the Ramon crater in the Negev desert of Israel.
All experiments are performed with the same vehicle setup. However, external sensor sources vary depending on the environment.

\subsection{University of Klagenfurt}
\label{sec:environment_uni}

The location at the University of Klagenfurt is used for two types of experiments.
First, the previously mentioned indoor datasets in a controlled environment for the initial tests of algorithms with the flight platform.
This scenario makes use of the "Dronehall", a motion capture environment with an area of \SI{150}{\meter\squared} and \SI{10}{\meter} in height. This area features $37$~cameras and provides millimeter and sub-degree accuracy.
Experiments in this environment also provide fiducial markers on the ground, visible by the navigation camera, and three \ac{UWB} modules placed at different heights. Light conditions for these experiments are constant.
The second type of experiment is the outdoor to indoor transition scenario. These experiments make use of the surrounding outdoor area and the indoor area of the Dronehall.
This experiment also made use of fiducial marker and \ac{UWB} modules but in a different configuration compared to the indoor experiments.
The environmental information which applies to both experiments is summarized by Table~\ref{tab:env_info_dronehall}.

\begin{table}[t]
    \small
    \ra{0.8}
    \caption{\ac{GNSS} coordinates and magnetic variation at the University of Klagenfurt test site - Austria, using the WMM 2019-2024 model \cite{noaa2019,Chulliat2020}}%
    \label{tab:env_info_dronehall}
    \begin{center}
        \begin{tabular}{ l| l }
        \toprule 
        \textbf{Location:} & Klagenfurt am Wörthersee - Austria \\
        Latitude: & $\SI{46}{\degree}$ $36'$ $48.813444"$ N \\
        Longitude: & $\SI{14}{\degree}$ $15'$ $44.59788"$ E \\
        Altitude(GPS): & $\SI{489.494}{\meter}$ \\
        \midrule
        \textbf{Magnetic field} & \\
        Date/Model: & 28. Feb. 2021, WMM 2019-2024 \\
        Declination:& $\SI{4}{\degree}$ $11'$ $25.8"$, positive (east) \\ 
        Inclination: & $\SI{63}{\degree}$ $7'$ $12.72"$, positive (down)\\ 
        Field strength: & $\SI{48302.7}{\nano\tesla}$ \\
        \bottomrule
        \end{tabular}
    \end{center}
\end{table}

\subsubsection{Motion Capture - Indoor}
\begin{figure}[htb]
    \centering
    \includegraphics[width=1.0\linewidth, trim=1.8cm 1.5cm 1.8cm 1.8cm, clip]{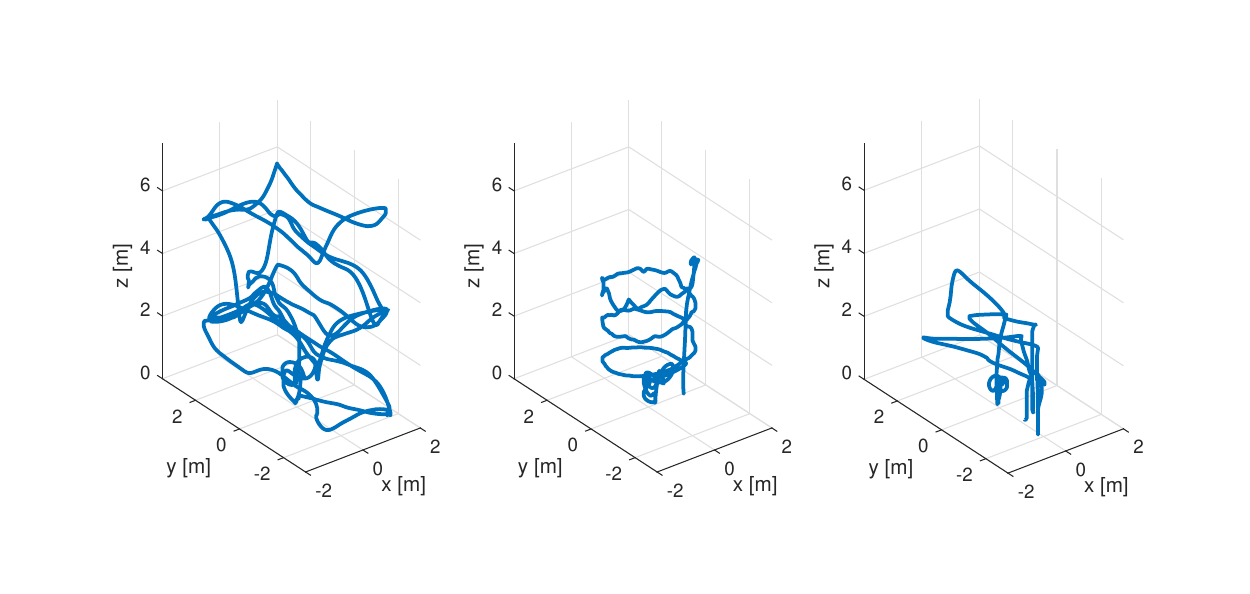}
    \caption{Indoor motion capture flight patterns. \textbf{Left:}~Ascending square trajectory, \textbf{Middle:}~Upwards spiral, \textbf{Right:}~Translations with pick and place elements.}
    \label{fig:indoor_flight_patterns}
\end{figure}

The indoor datasets cover three patterns at moderate speed:~An upward spiral movement, a square trajectory, and a pick-and-place scenario in which the vehicle moves to multiple locations and performs a touch-down and take-off maneuver.
The three patterns are visualized by Figure~\ref{fig:indoor_flight_patterns}. The pick and place scenarios have impulses in linear acceleration during the touch-down segments.
The pick and place scenario also features \ac{UWB} measurements for triangulation.

\subsubsection{Outdoor to Indoor Transition}
\begin{figure}[tb]
    \centering
    \includegraphics[width=1.0\linewidth]{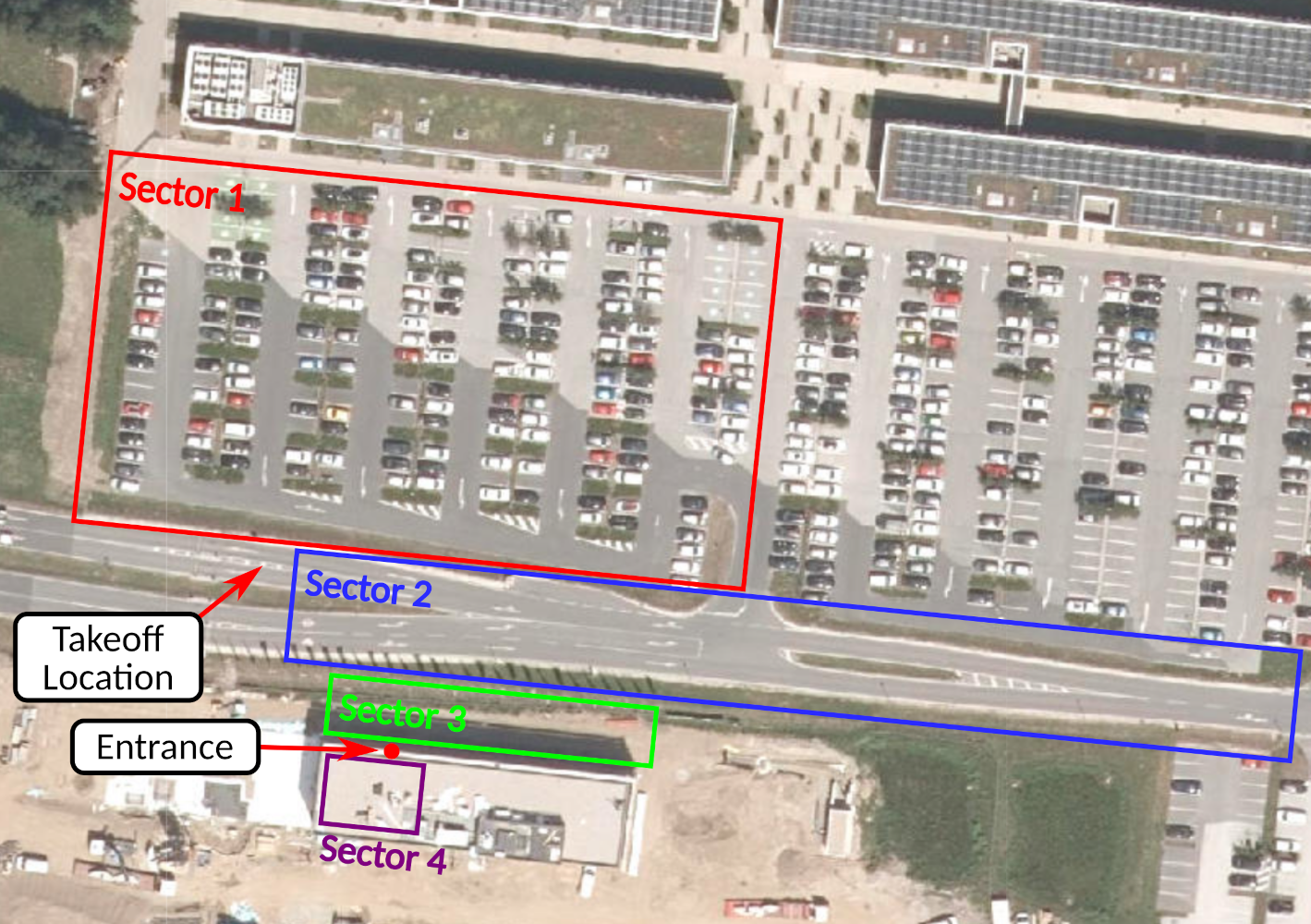}
    \caption{Location map and flight sectors for the Klagenfurt Dronehall indoor - outdoor transition experiments \protect\footnotemark}
    \label{fig:experiment_map_transition}
\end{figure}

\begin{figure}[tb]
    \centering
    \includegraphics[width=1.0\linewidth, trim=3.5cm 0.2cm 3.5cm 1cm, clip]{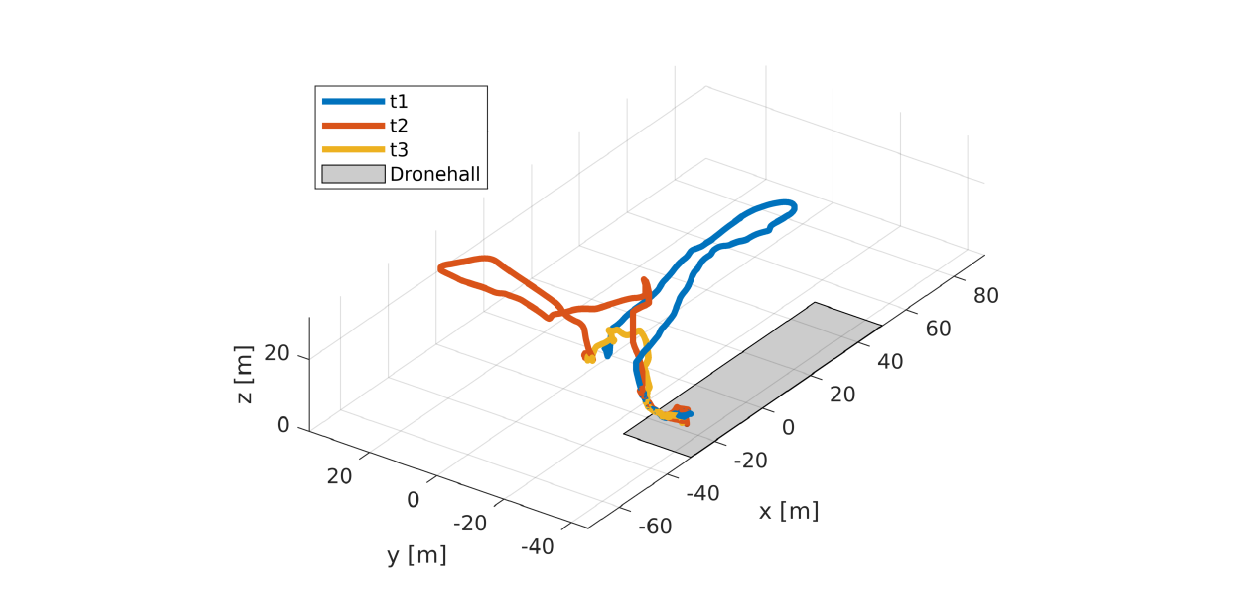}
    \caption{Trajectory and overlays for the outdoor to indoor transition datasets. The shaded area marks the indoor area.}
    \label{tab:transition_trajectory_overlays}
\end{figure}

The outdoor to indoor transition experiments are performed in the surrounding outdoor area, and the indoor area of the Dronehall.
This location made use of the fiducial marker setup as described in Section~\ref{sec:fiducial_marker}.
The transition datasets provide a variety of flight patterns. All datasets start at approximately the same location.
Figure~\ref{fig:experiment_map_transition} illustrates individual flight sectors and Figure~\ref{tab:transition_trajectory_overlays} shows the recorded trajectories and map overlays.

\footnotetext{Map source: Land Kaernten - \href{https://kagis.ktn.gv.at}{KAGIS}}

\begin{figure}[tb]
    \centering
    \includegraphics[width=1.0\linewidth, trim=1cm 0cm 0.8cm 0.8cm, clip]{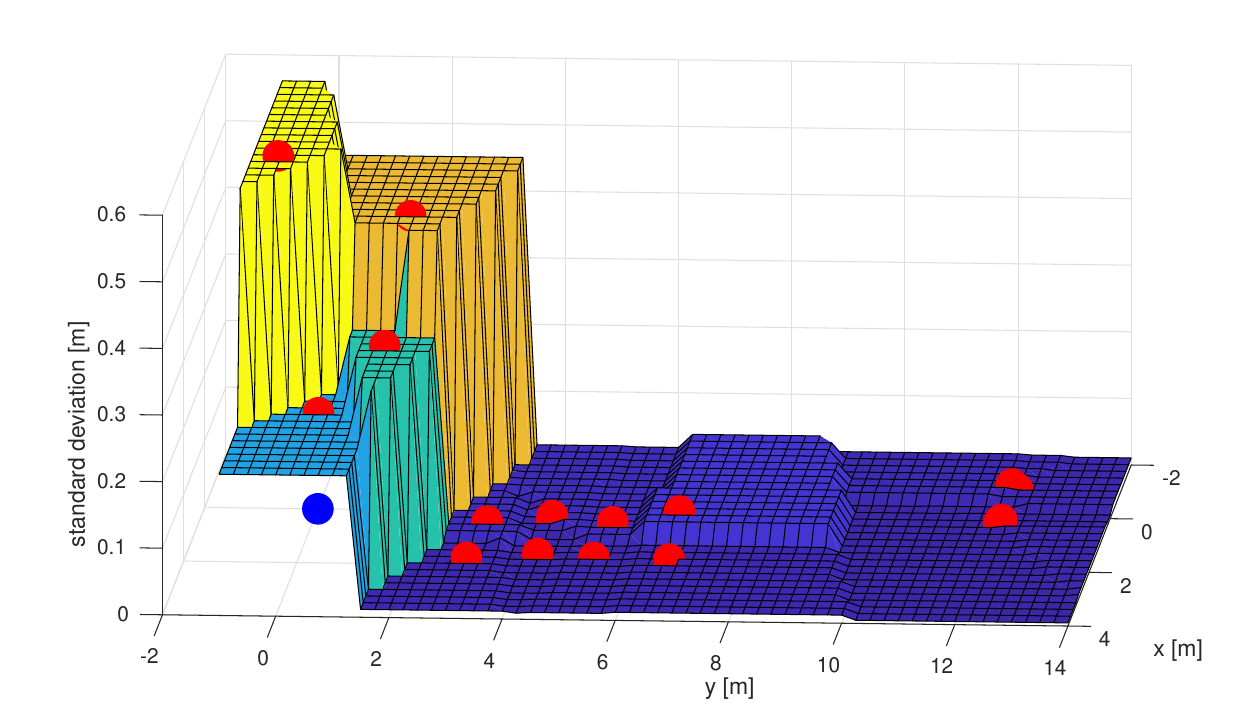}
    \caption{This map illustrates the \ac{RTK} \ac{GNSS} accuracy in close proximity to the Dronehall building. Red dots show the XY position at which the measurements are taken, and the z-axis shows the sample standard deviation of different static measurement sequences with more than $2000$~position measurements each.
    The surface shows the \textit{nearest interpolation} of the data points, and the blue dot indicates the location of the entrance. This shows that the position accuracy towards the building degrades as expected. It also shows that the accuracy is sufficient between four and eight meters, which is the area where the fiducial markers are placed for ground~truth during the transition phase.}
    \label{fig:gnss_accuracy_map}
\end{figure}

Sectors~1 and~2 are the main outdoor areas of the science park of the university. Sector~3 is the transition area in which the \ac{GNSS} quality starts to degrade based on the altitude of the vehicle.
The lower the position of the vehicle, the higher the \ac{GNSS} signal occlusion due to the building.
Figure~\ref{fig:gnss_accuracy_map} further illustrates the \ac{GNSS} signal degradation as the vehicle moves closer to the building at ground level.
Sector~4 represents the indoor location of the Dronehall.
The process of gaining and losing the \ac{RTK} fix of the \ac{GNSS} modules shows a hysteresis effect. Once the observation of satellites is lost after a longer period, regaining the same accuracy is not instantaneous and requires significant time.

\begin{figure}[tb]
    \centering
    \includegraphics[width=1.0\linewidth, trim=1.8cm 0.1cm 1.8cm 0.1cm, clip]{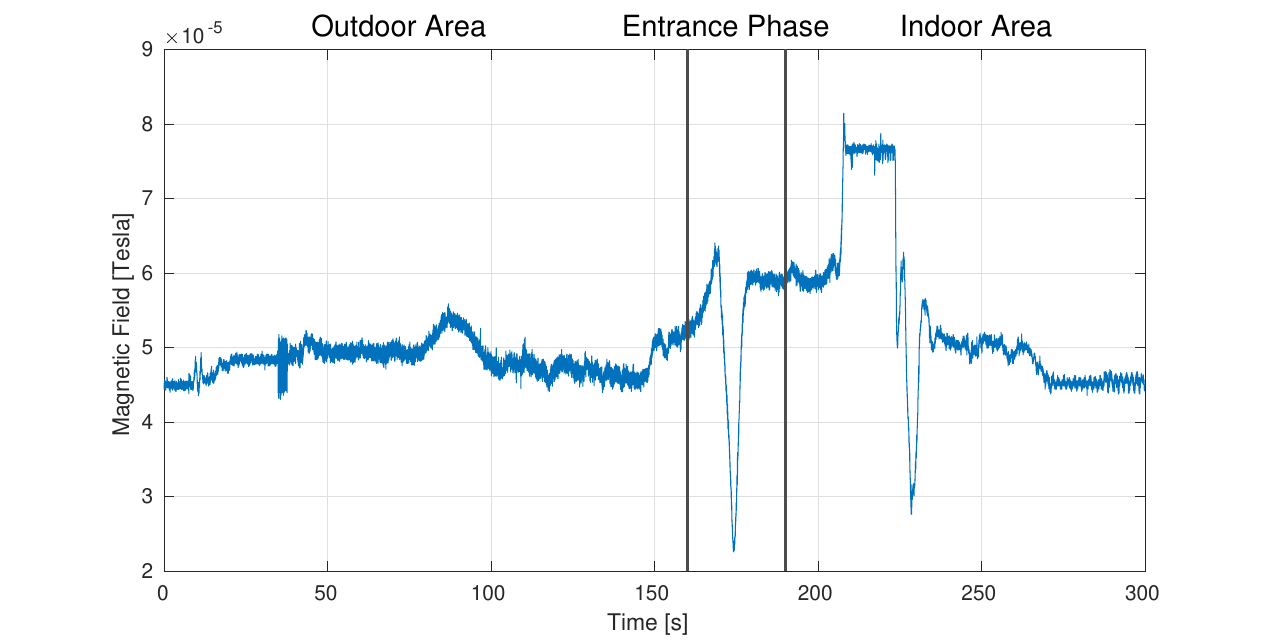}
    \caption{This graph shows the change of the magnetic field during the transition phase from outdoor to indoor. Such real-world effects can either be treated as outliers or as additional information to adapt to changing magnetic field as described by~\cite{Brommer2020_mag}.}
    \label{fig:mag_indoor_interference}
\end{figure}

Datasets were recorded throughout the day, which resulted in varying light conditions between recordings.
The lighting conditions for the transition datasets are especially challenging.
Sectors~1 and~2 provide natural light and possible reflections from the ground varying based on the height and rotation of the vehicle.
Sector~3 has shadow casting because of the nearby building, and Sector~4 provides artificial light after transitioning to the indoor location.
As described in Section~\ref{sec:camera}, the exposure time of the navigation camera was fixed and as low as possible to prevent motion blur. The gain was set to "auto" with a reference value that provides the best result for the changing light conditions.
The majority of trajectories perform a square over Sectors~1 and~2 before transitioning to the indoor location, starting at a high altitude of \SI{20}{\meter}.
One dedicated dataset also provides a constant velocity segment along Sector~2, to expose possible unobservability for VIO algorithms.
A subsection of datasets also shows lens flare effects for the stereo camera when facing the sun at high altitudes.

Another real-world effect is the change of the magnetic field strength when transitioning from indoor to outdoor due to the influence of the building structure.
Figure~\ref{fig:mag_indoor_interference} shows the change of the magnetic field strength during one transition.
Such real-world effects are desired and highly relevant for the development of real-world applications. The presented work will enable the development and deployment of robust solutions to work with such effects.

\subsection{Model Airfield Klagenfurt}
\label{sec:environment_airport}
The location at the model airfield in Klagenfurt, with the environmental information summarized by Table~\ref{tab:env_info_airport}, provides a purely outdoor dataset with partially visible fiducial marker and \ac{UWB} modules.
These data sequences focus on a simplified scenario for visual-inertial odometry and \ac{UWB} triangulation. The environment mainly provides a planar grass field with a partial agricultural area. Similar to the outdoor-indoor transition setup, the three \ac{UWB} modules are mounted on fiducial marker boards to calculate a global reference.
However, instead of using a motion capture system, the GNSS-based ground~truth is used to express the location of the board and the \ac{UWB} module position consecutively.

\begin{table}[htb]
    \small
    \ra{0.8}
    \caption{\ac{GNSS} coordinates and magnetic variation at the Klagenfurt model airfield test site - Austria}%
    \label{tab:env_info_airport}
    \begin{center}
        \begin{tabular}{ l| l }
        \toprule 
        \textbf{Location:} & Klagenfurt am Wörthersee - Austria \\
        Latitude: & $\SI{46}{\degree}$ $36'$ $24.7226364"$ N \\
        Longitude: & $\SI{14}{\degree}$ $16'$ $44.8363164"$ E \\
        Altitude(GPS): & $\SI{484.017}{\meter}$ \\
        \midrule
        \textbf{Magnetic field} & \\
        Date/Model: & 05. Feb. 2021, WMM 2019-2024 \\
        Declination:& $\SI{4}{\degree}$ $11'$ $3.12"$, positive (east) \\ 
        Inclination: & $\SI{63}{\degree}$ $6'$ $48.96"$, positive (down) \\ 
        Field strength: & $\SI{483298.5}{\nano\tesla}$ \\
        \bottomrule
        \end{tabular}
    \end{center}
\end{table}

\subsection{Mars Analog Desert}
\label{sec:environment_mars}
The desert location at the Ramon crater in Israel was used for the Mars analog simulation AMADEE20\footnote{\href{https://oewf.org/amadee-20/}{OeWF - AMADEE20}} and was the experiment area for the Mars analog Helicopter experiments.
The experiments were performed in a wider area, but information on a general environmental reference point is shown by Table~\ref{tab:env_info_israel}.

\begin{figure}[tb]
    \centering
    \includegraphics[width=1.0\linewidth]{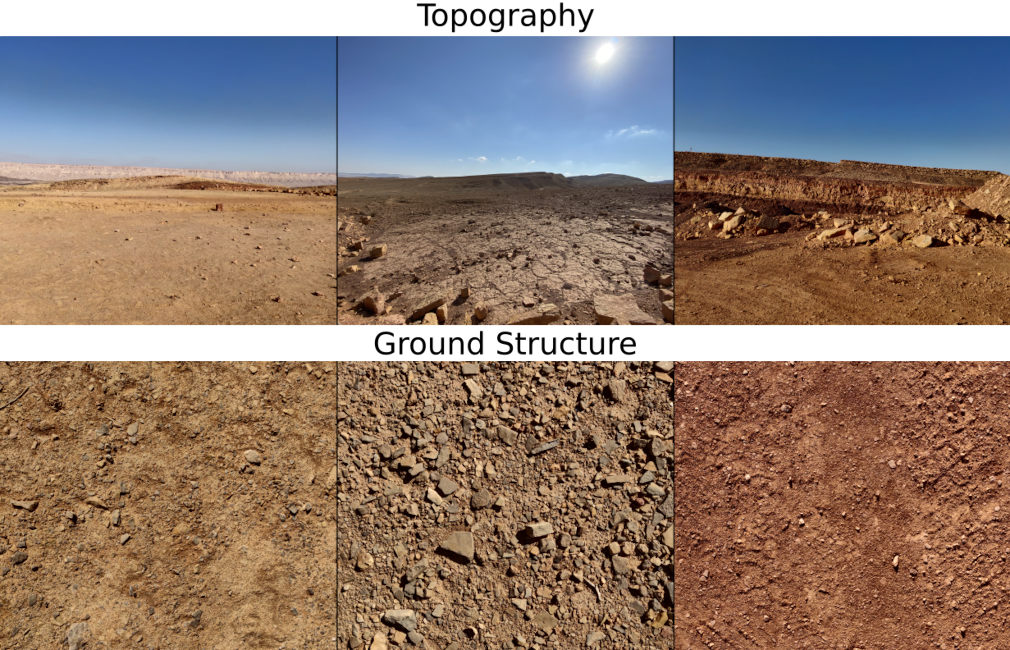}
    \caption{Externally recorded images as a reference for terrain structures. The top row shows the topography, and the bottom row shows a close-up of the ground structure. \textbf{Left:}~Sand patches and no terrain, \textbf{Middle:}~Stone paved medium terrain, \textbf{Right:}~High terrain elevation and sandy ground structure.}
    \label{fig:analog_diverse_ground_structure}
\end{figure}

\begin{figure}[tb]
    \centering
    \includegraphics[width=1.0\linewidth]{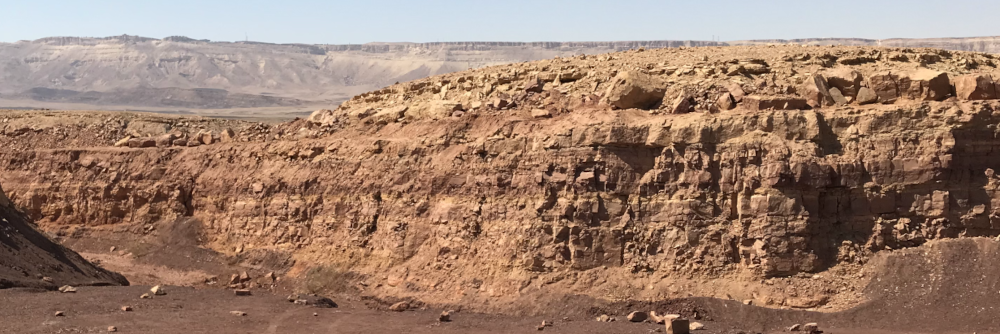}
    \caption{Representative image of a crater wall in the Negev desert. This location, among others was used to record data for a possible crater wall mapping task.
    The image was taken with an external device.}
    \label{fig:crater_wall_example}
\end{figure}

\begin{figure*}[!htbp]
    \centering
    \includegraphics[width=1\linewidth, trim=0cm 0cm 0cm 0cm, clip]{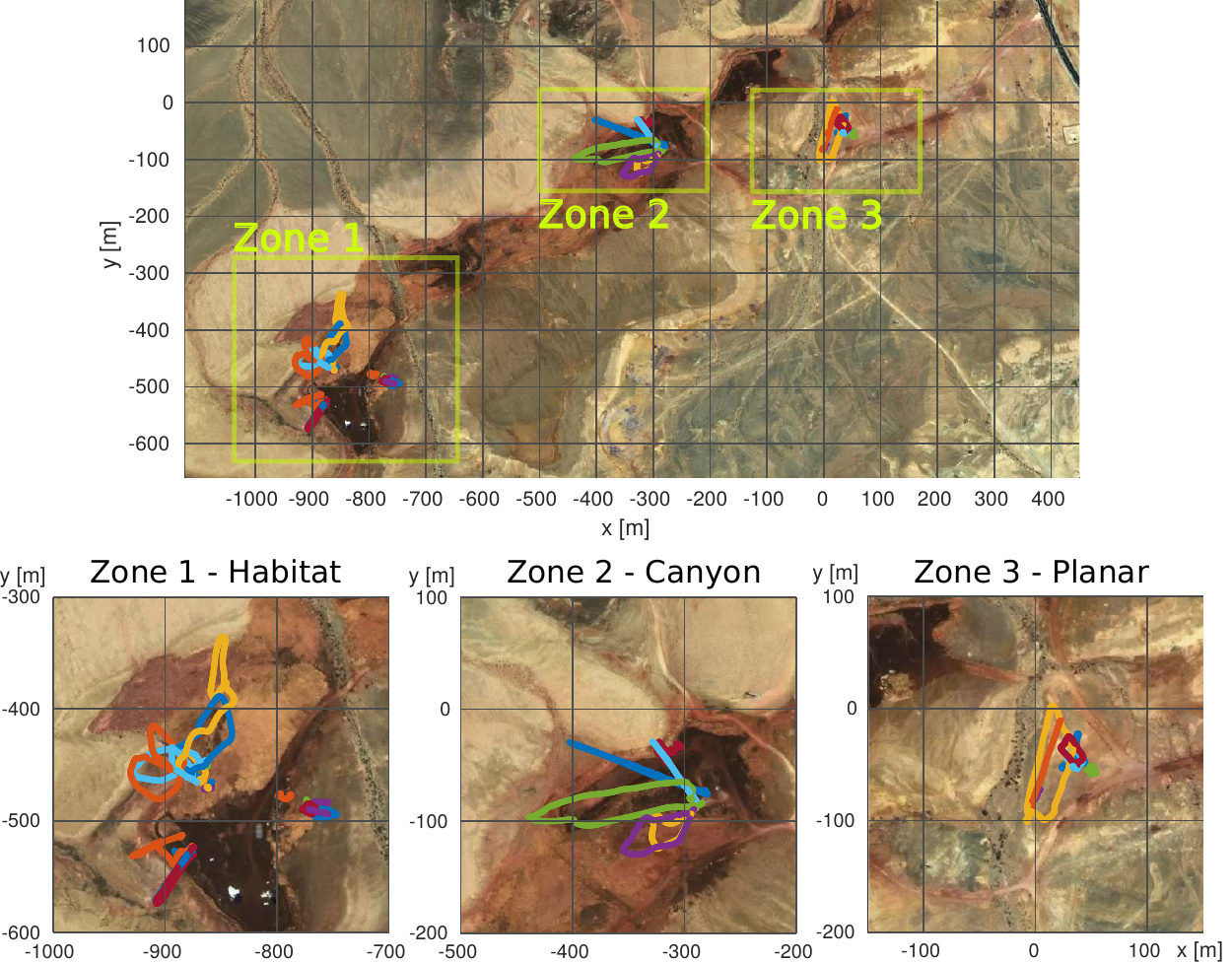}
    \caption{Trajectory and map overlays for the Mars analog datasets. Zone~1 shows the habitat in a small valley where a part of the cliff mapping data was recorded. Zone~2 shows a steep and lengthy canyon in which cliff mapping data and data with high elevation changes were acquired. Zone~3 was a planar-only region that was also used for the science close-up maneuvers. - Map \copyright OpenStreetMap contributors}
    \label{fig:desert_trajectory_overlays}
\end{figure*}

While the recorded sequences can be used as a multi-sensor setup without relation to off-world experiments, we want to highlight the specific high-level aspects regarding the Mars analog datasets concerning the environment, the sensor setup, and the design of the experiments.
The sensor setup of the vehicle includes an \ac{LRF} co-mounted with the navigation camera to resemble the specific sensor setup of the Mars helicopter Ingenuity.

The properties of the environment, such as the terrain topographies, rock distribution, granularity, coloring of the sediment structures, and other environmental parameters, are deemed to be similar to a Mars environment, according to the Austrian Space Forum \cite{Gromer2022}.
The datasets cover scenarios for different surface properties ranging from flat and sand-only surfaces to flat and stone-paved patches, as illustrated by Figure~\ref{fig:analog_diverse_ground_structure}. Special attention was also given to terrain structures with high elevation, such as small canyons, craters, and crater walls (see Fig.~\ref{fig:crater_wall_example}).
This environment does not feature external synthetic visual cues or sensors such as \ac{UWB} modules. The only external component was the \ac{RTK} \ac{GNSS} base station.

\begin{table}[htb]
    \small
    \ra{0.8}
    \caption{\ac{GNSS} coordinates and magnetic variation at the Ramon crater, Negev desert test site - Israel}%
    \label{tab:env_info_israel}
    \begin{center}
        \begin{tabular}{ l| l }
        \toprule 
        \textbf{Location:} & Negev Desert, Ramon Crater - Israel \\
        Latitude: & $\SI{30}{\degree}$ $35'$ $59.74384"$ N \\
        Longitude: & $\SI{34}{\degree}$ $52'$ $2.30880"$ E \\
        Altitude(GPS): & $\SI{526.594}{\meter}$ \\
        \midrule
        \textbf{Magnetic field} & \\
        Date/Model: & 08. Oct. 2021, WMM 2019-2024\\
        Declination:& $\SI{4}{\degree}$ $46'$ $21"$, positive (east) \\ 
        Inclination: & $\SI{46}{\degree}$ $16'$ $52.68"$, positive (down) \\ 
        Field strength: & $\SI{44486.3}{\nano\tesla}$ \\
        \bottomrule
        \end{tabular}
    \end{center}
\end{table}

Experiments were designed to cover Mars analog relevant scientific flight tasks and difficulties. Thus, experiments include science close-up flights with transitions from intermediate altitude towards an object such as a rock, a semi-circular scan of the object, and consecutive raises to the initial altitude.
Another experiment for the acquisition of scientific data is scans of crater walls. The Mars helicopter is predestined for such tasks because rovers were not able to record imagery of sediment layers at close proximity and high elevation.
Thus, datasets~9, 13, and 14 provide longer horizontal traverses along crater walls with the 3D~camera facing the wall. Both experiment types can be used for mapping and 3D reconstruction.
In addition to the camera-based information, we also performed experiments that pose challenges because of the flight patterns and the environment. These experiments include higher speed traverse at high altitude, constant velocity segments to challenge observability aspects of localization algorithms, and crater wall fly-overs for abrupt and relative elevation changes.

Figure~\ref{fig:desert_trajectory_overlays} shows the majority of flight patterns performed in the Ramon Crater.
The flights were performed in three distinct zones.
Zone one is the location of the habitat and had semi-structured ground and terrain with the exception of a crater wall, which was used for trajectories with high relative altitude changes and vertical movement along the crater wall for possible mapping and 3D reconstruction.
Zone two has terrain with step elevation changes in the form of a canyon. This zone was used for lateral crater wall recordings and long distances with higher altitude flights.
The ground structure in zone three was a planar ground with low texture and was used for low-altitude science close-ups and multiple lateral traverse flights.

\subsection{Datasets}

%
\begin{table*}[htb]
    \setlength\tabcolsep{3pt}
    \caption{Table of main datasets. The local locations for the experiments are shown in Figures~\ref{fig:experiment_map_transition} and~\ref{fig:desert_trajectory_overlays}.}%
    \label{tab:dataset_list}
    \begin{center}
    {
      \resizebox{\textwidth}{!}
      {%
        \small 
        
\begin{tabular}{|>{\centering}m{0.5cm}>{\centering}m{1cm}>{\raggedright}m{3.5cm}>{\raggedright}m{4.5cm}>{\centering}p{1cm}>{\centering}p{1cm}>{\centering}p{1cm}>{\centering}p{1.5cm}|}
\hline 
No. & Test

Site & \multicolumn{1}{>{\centering}m{3.5cm}}{Pattern} & \multicolumn{1}{>{\centering}m{4.5cm}}{Difficulty} & \multicolumn{1}{>{\centering}m{1cm}}{Dist.

$[m]$} & \multicolumn{1}{>{\centering}m{1cm}}{max.

Vel.

$[m/s]$} & \multicolumn{1}{>{\centering}m{1cm}}{max.

Height 

$[m]$} & Aspects\tabularnewline
\hline 
\hline 
\multicolumn{8}{|l|}{\textbf{Dronehall Indoor}}\tabularnewline
\hline 
1 &  & Ascending square pattern & \multirow{2}{4.5cm}{None, used for baseline testing} & 60 & 1.0 & 7 & \tabularnewline
\cline{1-3} \cline{2-3} \cline{3-3} \cline{5-8} \cline{6-8} \cline{7-8} \cline{8-8} 
2 &  & Upward spiral &  & 25 & 0.3 & 5 & \tabularnewline
\hline 
3 &  & Pick and place & Low, IMU impulses on intermediate landing & 42 & 1.0 & 3 & with UWB\tabularnewline
\hline 
\hline 
\multicolumn{8}{|l|}{\textbf{Outdoor to indoor transition: }\textit{Changing light conditions,
magnetic field, and sensor availability. }}\tabularnewline
\multicolumn{8}{|l|}{\textit{~~No constant absolute sensor information besides ground
truth.}}\tabularnewline
\hline 
1 &  & Constant velocity

along sector 2 $\rightarrow$ sector 4 & High, observability aspects to VIO & 80 & 2.0 & 20 & with UWB\tabularnewline
\hline 
2 &  & Square in sector 1

$\rightarrow$ sector 2 $\rightarrow$ sector 4 & High & 120 & 2.0 & 25 & with UWB\tabularnewline
\hline 
3 &  & Square in sector 1

$\rightarrow$ sector 4 & High, natural lense flair on stereo cam. & 180 & 2.5 & 32 & with UWB\tabularnewline
\hline 
\hline 
\multicolumn{8}{|l|}{\textbf{Model Airfield}}\tabularnewline
\hline 
1 &  & Square pattern & Low, simplified outdoor only trajectory & 260 & 6.0 & 25 & with UWB\tabularnewline
\hline 
\hline 
\multicolumn{8}{|l|}{\textbf{Mars Analog Desert: }\textit{Semi structured ground and terrain}}\tabularnewline
\hline 
1 & Zone 1 & Circular pattern & Low, high structured ground & 65 & 2.3 & 6 & \tabularnewline
\hline 
2 & Zone 1 & Circular pattern & Low, medium structured ground & 98 & 3.5 & 19 & \tabularnewline
\hline 
3 & Zone 1 & Circular pattern & Low, high structured ground & 190 & 5.0 & 20 & \tabularnewline
\hline 
4 & Zone 1 & Traverse forth and back & Medium, lower structured ground & 145 & 4.5 & 25 & \tabularnewline
\hline 
5 & Zone 1 & Traverse forth and back & Medium, lower structured ground & 27 & 2.0 & 9 & \tabularnewline
\hline 
6 & Zone 1 & Traverse forth and back & Low, no vision sensors & 48 & 2.0 & 13 & \tabularnewline
\hline 
7 & Zone 1 & Traverse forth and back & Low, high fine grained structured ground & 48 & 3.0 & 13 & \tabularnewline
\hline 
8 & Zone 1 & Cliff dive & High, high relative elevation changes & 80 & 4.0 & 10 & \tabularnewline
\hline 
\hline 
\multicolumn{8}{|l|}{\textbf{Mars Analog Desert: }\textit{Terrain with steep elevation
change}}\tabularnewline
\hline 
9 & Zone 2 & Traverse forth and back & Low, high structured ground & 70 & 4.0 & 15 & \tabularnewline
\hline 
10 & Zone 2 & Traverse forth and back & Low, high structured ground & 79 & 4.5 & 22 & \tabularnewline
\hline 
11 & Zone 2 & Traverse forth and back & Low, high structured ground & 158 & 6.0 & 40 & \tabularnewline
\hline 
12 & Zone 2 & Cliff dive & \multirow{2}{4.5cm}{High, high relative elevation changes} & 127 & 5.5 & 23 & \tabularnewline
\cline{1-3} \cline{2-3} \cline{3-3} \cline{5-8} \cline{6-8} \cline{7-8} \cline{8-8} 
13 & Zone 2 & Cliff dive &  & 60 & 2.0 & 12 & \tabularnewline
\hline 
14 & Zone 2 & Traverse, landing at new location & Medium, long distance,

if VIO only & 226 & 5.3 & 22 & \tabularnewline
\hline 
\hline 
\multicolumn{8}{|l|}{\textbf{Mars Analog Desert: }\textit{Low texture and planar ground}}\tabularnewline
\hline 
15 & Zone 3 & Science close-up & Medium, close proximity onjects & 90 & 6.0 & 2.5 & \tabularnewline
\hline 
16 & Zone 3 & Traverse landing at new location & Medium, high fine grained structured ground & 50 & 4.5 & 15 & \tabularnewline
\hline 
17 & Zone 3 & Traverse forth and back & High, lower structured ground & 162 & 5.0 & 15 & \tabularnewline
\hline 
18 & Zone 3 & Multiple takeoff and landings & \multirow{2}{4.5cm}{High, frequent elevation changes and dust occluded vision due to downwash} & 5 & 1.5 & 3.5 & \tabularnewline
\cline{1-3} \cline{2-3} \cline{3-3} \cline{5-8} \cline{6-8} \cline{7-8} \cline{8-8} 
19 & Zone 3 & Multiple takeoff and landings &  & 113 & 1.6 & 3 & \tabularnewline
\hline 
\hline 
\multicolumn{8}{|l|}{\textbf{Calibration data}}\tabularnewline
\hline 
\multicolumn{8}{|l|}{~~- Magnetometer intrinsic and extrinsic}\tabularnewline
\hline 
\multicolumn{8}{|l|}{~~- LRF + IR indicator (static)}\tabularnewline
\hline 
\multicolumn{8}{|l|}{~~- LRF + IMU (moving)}\tabularnewline
\hline 
\multicolumn{8}{|l|}{~~- All IMU 4h static}\tabularnewline
\hline 
\multicolumn{8}{|l|}{~~- Vehicle vibration data}\tabularnewline
\hline 
\multicolumn{8}{|l|}{~~- Navigation camera intrinsic and extrinsic (Klagenfurt experiments)}\tabularnewline
\hline 
\multicolumn{8}{|l|}{~~- Navigation camera intrinsic and extrinsic (Mars analog experiments)}\tabularnewline
\hline 
\multicolumn{8}{|l|}{~~- Stereo camera intrinsic and extrinsic (Klagenfurt experiments)}\tabularnewline
\hline 
\multicolumn{8}{|l|}{~~- Stereo camera intrinsic and extrinsic (Mars analog experiments)}\tabularnewline
\hline 
\end{tabular}
      }%
    }
    \end{center}
\end{table*}
%

This section describes the patterns and scenarios which were performed for each location.
In general, the majority of datasets have an initialization phase. This initialization phase consists of a smooth impulse in the direction of the IMU's \mbox{z-axis} and consecutive smooth excitations in 6~DoF.
This can be used for VIO algorithms requiring initial IMU excitement and for initial convergence.
However, since this feature is not desired in all cases, the copter is placed down to the ground after initialization, followed by a short steady pause before the actual flight maneuver. Thus, the initial phase can be cropped.
A list of the majority of recorded flights is shown in Table~\ref{tab:dataset_list}.
\FloatBarrier

\subsection{Vehicle Vibration Analysis}
\label{sec:vehicle_vibration_analysis}
It is well known that parasitic effects in measured inertial data can lead to drastic decreases in performance of classical state estimation methods if not accounted for. Apart from noise and biases of the inertial sensor itself, these also include vibrations of the vehicle that couple into the IMU at various frequencies. One major source of vibrations on UAVs are the motors. While it may be difficult to completely account for vibrations in classical state estimation approaches, it has been hypothesized that latent features in IMU data such as RPM or velocity dependent vibrations are beneficial for learning based methods \cite{Chen2018_ionet, Steinbrener2022_imupropagation}. In this section, we show that the motors introduce a resonant frequency in the IMU spectrum that is characteristic of their speed. This resonance is particularly pronounced in the inertial data obtained from the high-rate IMU.  

\begin{figure}[htb]
    \centering
    \includegraphics[width=1\linewidth, trim=1cm 0.0cm 1cm 0.3cm, clip]{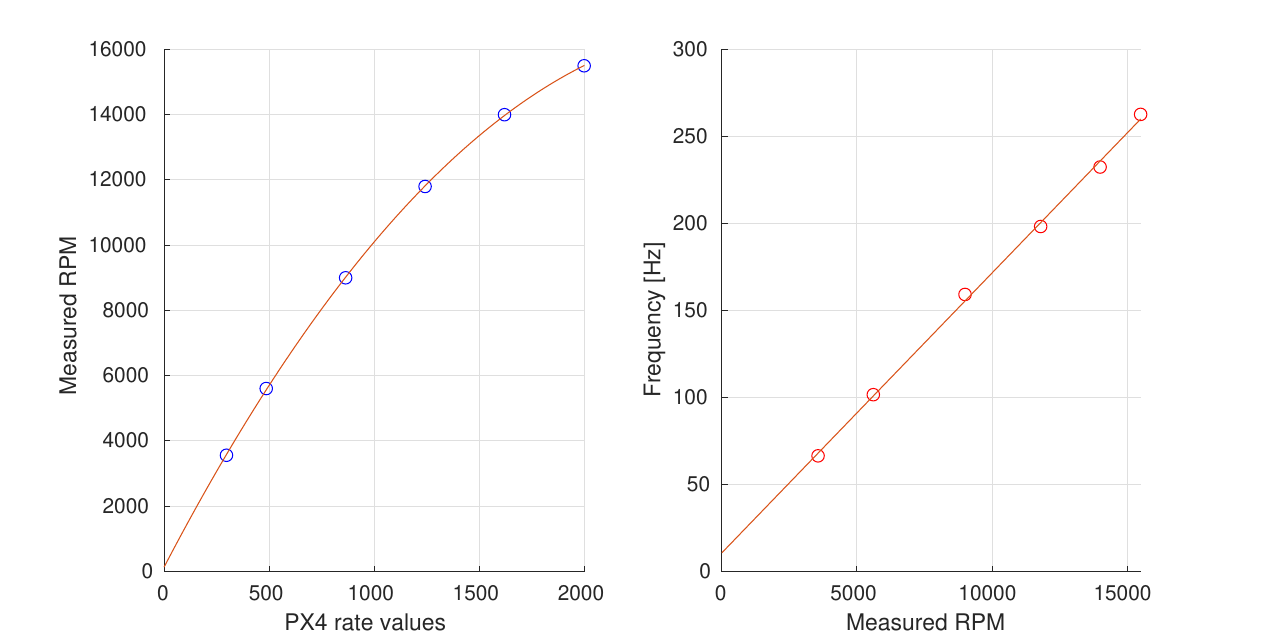}
    \caption{Calibration of PX4 rate values~\textbf{(left)}, and resonant frequencies as a function of motor RPM~\textbf{(right)}.}
    \label{fig:rpm-calib-resonances}
\end{figure}

During flight, the PX4 provides the rate values for each of the four motors at \SI{100}{\hertz}. The PX4 rate values range from $0$~to~$2000$ on a scale with arbitrary units. To calibrate these rate values to actual RPM values, the data obtained with the vibration measurement setup described in Sec.~\ref{sec:vibration_test_bench} for seven different motor speeds is used to derive an analytic relationship between PX4 rate values and measured RPM. The corresponding data is shown in Tab.~\ref{tab:nom-rpm-calib}, and plotted on the left hand size of Fig.~\ref{fig:rpm-calib-resonances} together with a quadratic polynomial with coefficients \mbox{$a_0=168.5541$}, \mbox{$a_1=12.1870$}, and \mbox{$a_2=-0.0023$}.

\begin{table}[htb]
    \small
    \ra{0.8}
    \caption{RPM Calibration}%
    \vspace{-0.4cm}
    \label{tab:nom-rpm-calib}
    \begin{center}
        \begin{tabular}{c| c| c }
        \toprule 
        Relative RPM & PX4 Rate Values & Measured RPM \\
        \midrule
        \SI{10}{\percent} & 297 & 3560 \\
        \SI{20}{\percent} & 486 & 5605 \\
        \SI{40}{\percent} & 864 & 9000 \\
        \SI{60}{\percent} & 1242 & 11800 \\
        \SI{80}{\percent} & 1620 & 14000 \\
        \SI{100}{\percent} & 1999 & 15500 \\
        \bottomrule
        \end{tabular}
    \end{center}
\end{table}

\begin{figure}[htb]
    \centering
    \includegraphics[width=1.0\linewidth]{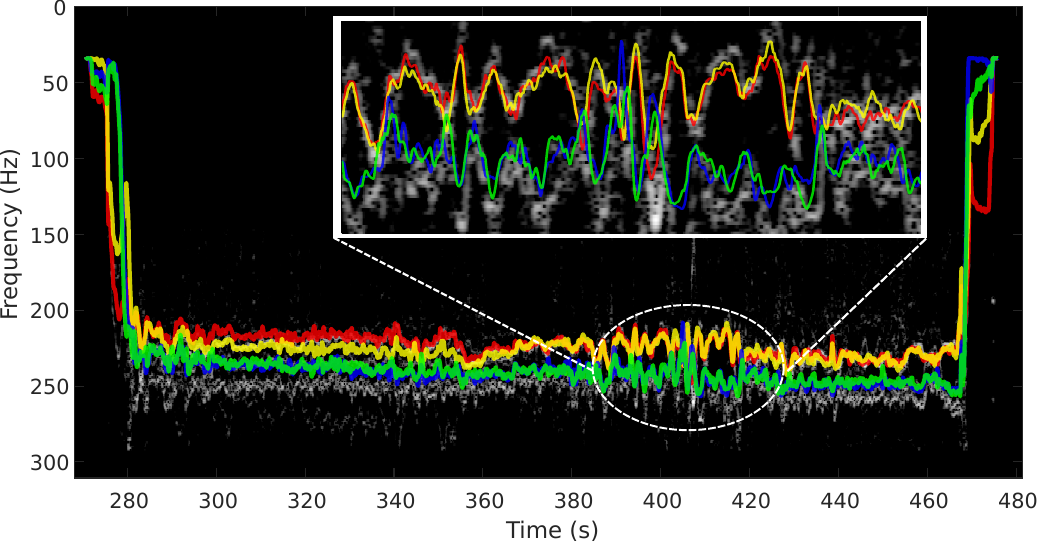}
    \caption{Prediction of spectral resonances based on PX4 rate values (color) and measured resonances (grey) for real-flight data.}
    \label{fig:vibration_spectrum_analysis}
\end{figure}

To analyze the frequencies of the resonances for the different motor speeds, we computed the averaged power spectral densities of the high-rate inertial data for each vibration test run using Matlab's \texttt{spectrogram} method, and extracted the frequency of the main peak using Matlab's \texttt{findpeaks} method. Plotting the true motor RPMs against the peak frequencies shows a linear relationship. The corresponding line fit with coefficients \mbox{$a_0=10.6666$} and \mbox{$a_1=0.0161$} is shown on the right side of Fig.~\ref{fig:rpm-calib-resonances} along with the data. Although this data was recorded in a very controlled experimental setup, a clear relationship between the resonance in the spectrogram and the PX4 motor rate values exists also for real-flight data. Figure~\ref{fig:vibration_spectrum_analysis} shows the spectrogram of the high-rate IMU for one of the flights in gray. Overlaid in color are the expected resonances computed from the reported PX4 rate values for each motor using the same linear fit parameters.
The inset shows a magnified portion of the whole spectrogram. As can be seen, the expected resonant frequencies correlate very well with the observed resonances. We attribute the remaining differences to experimental effects in the real-flight scenario. In particular, the actual RPMs may differ from the PX4 rate values. Nonetheless, these results indicate the presence of motor speed dependent resonances in the recorded inertial data that can be of interest in particular to the learning community.

\section{Ground Truth}
\label{sec:groung_truth}
This section describes the generation of 6~DoF ground~truth data for the outdoor datasets, sensor time calibration, as well as the alignment of the pose information for the outdoor to indoor transition datasets.
The tools which were used to generate the ground~truth will be open-sourced together with this dataset.

\subsection{Notation}
%
The notation of transformations used by this work is as follows.
$\rvar{A}{B}$ is the rotation of frame~$B$ expressed in frame~$A$. 
$\vvar{p}[A][B][A] \equiv \vvar{p}[A][B]$ is the translation from frame~$A$~to~$B$ expressed in frame~$A$. As an example $\vvar{p}[A][B][C] = \rvar{C}{A}~\vvar{p}[A][B][A]$.

Respective homogeneous transformations are defined by
\begin{align}
&* : \mathbb{R}^{4\times4} \times \mathbb{R}^3 \rightarrow \mathbb{R}^3,  \nonumber\\
\text{with} ~
\vvar{T}[A][B] &= \bracketMatrixstack{
       \rvar{A}{B} & ~\vvar{t}[A][B], \\
       0_{1\times3} & 1} \in \mathbb{R}^{4\times4}, \pvar{v}[B] \in \mathbb{R}^3, \nonumber\\
\vvar{T}[A][B] * \pvar{v}[B] &= \bracketMatrixstack{
       \rvar{A}{B} & ~\vvar{t}[A][B], \\
       0_{1\times3} & 1}
     * \vvar{v}[B] \mapsto \rvar{A}{B} \vvar{v}[B]+\vvar{t}[A][B]. 
\end{align}
\label{sec:notation}

\subsection{Vehicle 6~DoF Pose}
\label{sec:ground_truth_vehicle_pose}
As mentioned before, we chose to use individual raw measurements for the generation of the ground~truth data because of the high measurement accuracy. Using a recursive algorithm or a graph-based optimization for ground~truth data generation can cause biased results for comparisons against this ground~truth data. Original raw sensor data is provided to allow the use of different methods for the generation of ground~truth.

\begin{figure}[tb]
    \centering
    \includegraphics[width=1.0\linewidth, trim=0cm 0cm 0cm 0 , clip]{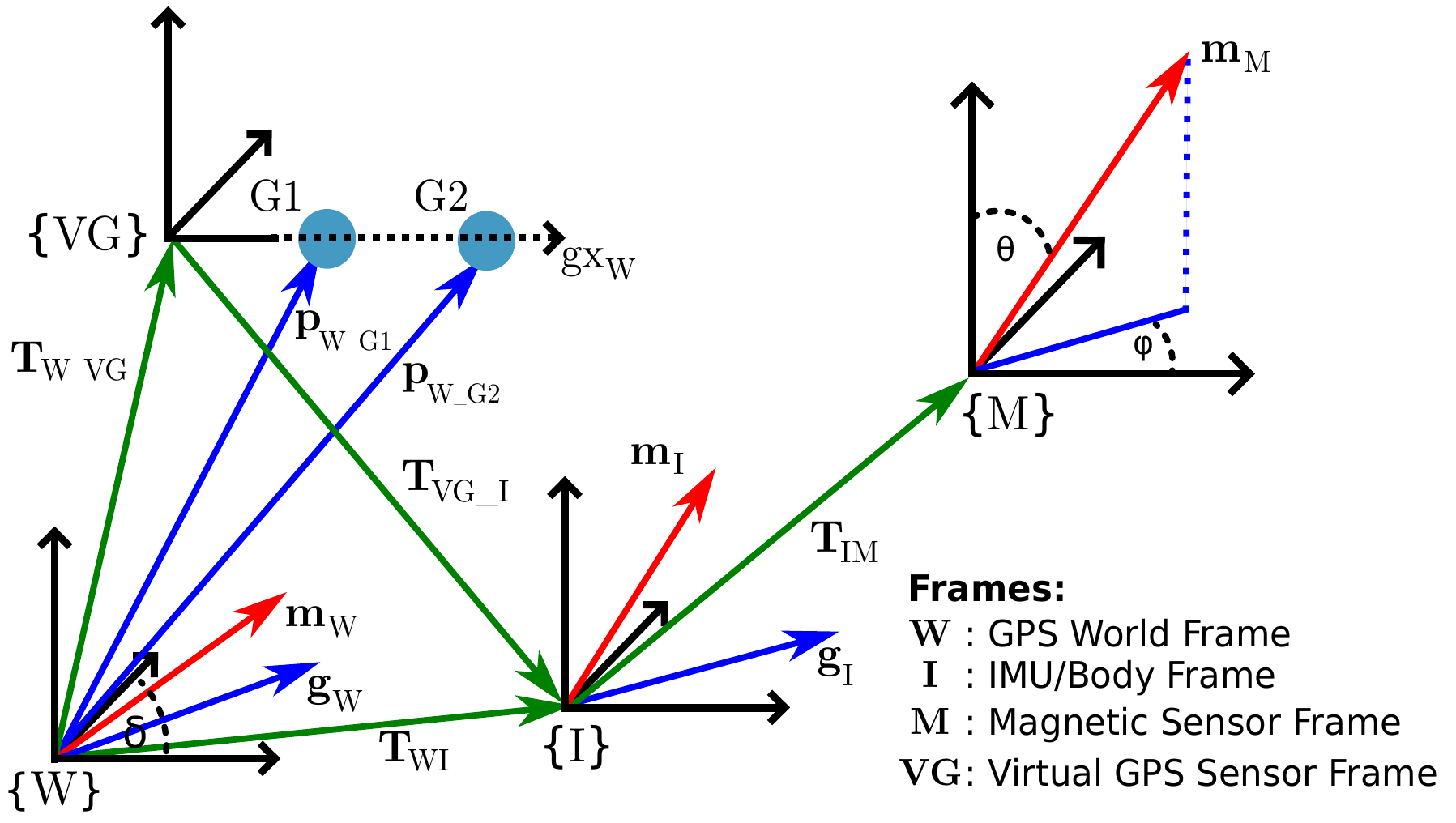}
    \caption{Ground~truth sensor frame setup. \ac{RTK} \ac{GNSS} and Magnetometer sensor-frames with respective transformations for the described 6~DoF ground~truth calculation.}
    \label{fig:ground_truth_sensor_fames}
\end{figure}

We aim to generate 6~DoF ground~truth information of an IMU-frame with respect to the \ac{GNSS} world-frame (ENU). The provided ground~truth for the vehicle is referenced to the PX4 IMU. This is done based on two \ac{GNSS} position measurements and a known calibration of the \ac{GNSS} antenna positions with respect to the IMU-frame.
The method relies on the generation of a virtual \ac{GNSS} sensor frame, which provides 6~DoF. The calculation of the pose for the IMU frame is then achieved by applying transformations of known calibration transformations.
Using the two position measurements and generating a virtual directional vector, this setup only provides rotational information in 2~DoF. Thus, we are using the measurements of the directional magnetic vector as support and to fix the ground~truth frame for a 3~DoF rotation. As a result, this method will lead to a full-frame definition in 3D with 3~DoF orientation and 3~DoF position. The reference frames and sensor information for the ground~truth method are shown in Figure~\ref{fig:ground_truth_sensor_fames}.
For the remainder of this document, the world-frame $\{W\}$ is equal to the \ac{GNSS} world-frame, following the East-North-Up~(ENU) convention \mbox{$\text{ENU} \rightarrow \text{XYZ}$}. The virtual \ac{GNSS} frame $\{VG\}$, IMU/Body frame $\{I\}$, and the frame of the magnetic-sensor $\{M\}$ are placed on a rigid body.
The \ac{GNSS} position measurements $\vvar{p}[W][G1]$ and $\vvar{p}[W][G2]$ are expressed with respect to the world frame.

The magnetic field $\pvar{m}[w]$ is assumed to be locally static with respect to the world-frame and the local magnetic variation, i.e. the inclination and declination, is known.
In the IMU frame, the magnetic variation depends on the rotation of the vehicle. This effect needs to be taken into account. Thus, a system that describes the rotation of the vehicle needs to use the following set of information: The magnetic vector expressed in the IMU and the world-frame $\pvar{m}[w]$ and $\pvar{m}[i]$, respectively, as well as the virtual \ac{GNSS} vectors in world and IMU frame $\pvar{g}[w]$ and $\pvar{g}[i]$, respectively. Since only direction is relevant, all vectors are normalized for this analysis. In addition, we can make use of the orthogonal vectors \mbox{$\pvar{c}[w] = \pvar{g}[w] \times \pvar{m}[w]$} and $\pvar{c}[i] = \pvar{g}[i] \times \pvar{m}[i]$ which renders a possible linear least squares problem full-rank. We present three different methods to analyze the resulting data. The naive approach is to solve the resulting linear least squares problem directly:
\begin{align}
    \setstackgap{L}{1.1\baselineskip}
    \fixTABwidth{T}
    \setstacktabbedgap{5pt}
       \mathbf{y} &= \mathbf{A}~\mathbf{x}\\
        \bracketMatrixstack{
       \pvar{g}[w]; & \pvar{m}[w]; & \pvar{c}[w]
    } &= \rvar{w}{i}
        \bracketMatrixstack{
        \pvar{g}[i]; & \pvar{m}[i]; & \pvar{c}[i]
    }
\end{align}
However, an additional adaptation is needed because the result is not guaranteed to be a rotation matrix. Another approach is given by the following system, which allows a non-linear optimization on the tangent space and ensures that the result is a rotation matrix:
\begin{equation}
    \setstackgap{L}{1.1\baselineskip}
    \fixTABwidth{T}
    \setstacktabbedgap{5pt}
        \bracketMatrixstack{
       \pvar{g}[w] \\
       \pvar{m}[w] \\
       \pvar{c}[w]
    } = 
    \bracketMatrixstack{
       e^{\lfloor\boldsymbol{\omega}\rfloor} & 0 & 0 \\
       0 & e^{\lfloor\boldsymbol{\omega}\rfloor} & 0 \\
       0 & 0 & e^{\lfloor\boldsymbol{\omega}\rfloor}
    }
        \bracketMatrixstack{
       \pvar{g}[i] \\
       \pvar{m}[i] \\
       \pvar{c}[i]
    }
\end{equation}
With Jacobian:
\begin{equation}
    \setstackgap{L}{1.1\baselineskip}
    \fixTABwidth{T}
    \setstacktabbedgap{5pt}
    \frac{\partial}{\partial \boldsymbol{\omega}}
    h(\boldsymbol{\omega}) = 
    \bracketMatrixstack{
       -e^{\lfloor\boldsymbol{\omega}\rfloor}~\lfloor \pvar{g}[i] \rfloor; & -e^{\lfloor\boldsymbol{\omega}\rfloor}~\lfloor \pvar{m}[i] \rfloor; & -e^{\lfloor\boldsymbol{\omega}\rfloor}~\lfloor \pvar{c}[i] \rfloor
    }
    \label{eq:coherent_cov}
\end{equation}
Finally, our preferred method because of its simplicity, forms a least squares problem according to Wahba's problem \cite{Wahba1965}. This approach ensures a rotation matrix and allows to introduce weights for individual inputs.
For the present application, it is of interest to associate a higher weight to the \ac{GNSS}-based information because it is less likely to be disturbed by the environment in contrast to the magnetometer.
The magnetometer measurements only serve the purpose of supporting the third degree of freedom for the orientation information. The weight $\alpha$ for the \ac{GNSS} vector is chosen heuristically and is set to~50:1.
\begin{align}
\mathbf{A} &= \alpha~\pvar{g}[w]~\pvar{g}[i]^\text{T} + \pvar{m}[w]~\pvar{m}[i]^\text{T} + \pvar{c}[w]~\pvar{c}[i]^\text{T}\\
\mathbf{A} &= \mathbf{U}\mathbf{S}\mathbf{V}^\text{T}\\
\mathbf{R} &= \mathbf{U}\mathbf{M}\mathbf{V}^\text{T}\\
\text{with~} \mathbf{M} &= \text{diag}\left(\bracketMatrixstack{1; & 1; & \text{det}(\mathbf{U})\text{det}(\mathbf{V})
    }\right)
\end{align}
The position of the IMU expressed in the world-frame, given that \ac{GNSS} point measurement $G2$ is the origin of the virtual \ac{GNSS}-frame, is then calculated based on
\begin{align}
    \vvar{p}[W][G2] &= \vvar{T}[W][I]*\vvar{p}[I][G2]\\
    &= \vvar{T}[W][VG]~\vvar{T}[VG][I]*\vvar{p}[I][G2]\\
    \Rightarrow \vvar{p}[W][I] &= \vvar{p}[W][G2] + \rvar{W}{VG} ~ \rvar{VG}{I} ~ (-\vvar{p}[I][G2]).
\end{align}
The three methods for rotation estimation described above are provided with the dataset tools and can be chosen accordingly.

In this context, it is also important to note that the rotation can be incorrect if the magnetometer provides incorrect information. This can be the case where the vehicle is picked up at the beginning of an experiment and the magnetic field is disturbed by the person of the field crew. The increased weight on the \ac{GNSS} vector when calculating the 3~DoF rotation reduces this effect during this time, and such disturbance does not occur during the main phase of an experiment.

\subsection{Sensor Time Calibration}
\label{sec:gt_time_sync}
The synchronization for the computation platforms of the vehicle are described in Section~\ref{sec:module_synchronization}, however, internal system delays can still lead to minor time shifts of the sensor data.
Thus, all sensors that are required for the generation of ground~truth data are additionally time-synchronized in post-processing.
This includes the two \ac{RTK} \ac{GNSS} signals, the PX4 IMU and magnetometer, as well as the motion capture system.

To synchronize the two \ac{GNSS} signals, we first calculate their velocity and use their convolution to find the time offset which provides the highest overlap of the two signals $\dot{\mathbf{p}}_\mathsf{g1}$ and $\dot{\mathbf{p}}_\mathsf{g2}$.
Using the synchronized \ac{GNSS} information, the virtual orientation vector for the two \ac{GNSS} measurements expressed in the world frame is built (see the calculation of $VG$ in Section~\ref{sec:ground_truth_vehicle_pose}). 
In order to synchronize the magnetometer data to the virtual \ac{GNSS} vector, the time derivative of the horizontal projection for $VG$ and the yaw rotation component measured by the magnetometer are used.

After this step, the time derivative of the rotational ground~truth for the outdoor pose information and the motion capture pose is used for the synchronization towards the angular velocity of the PX4 IMU.
This provides all sensor time-offsets to generate accurate ground~truth information for the vehicle.

\subsection{Transition Segment Alignment}
\begin{figure}[bt]
    \centering
    \includegraphics[width=1.0\linewidth, trim=1.3cm 0.2cm 1.3cm 0.3cm, clip] {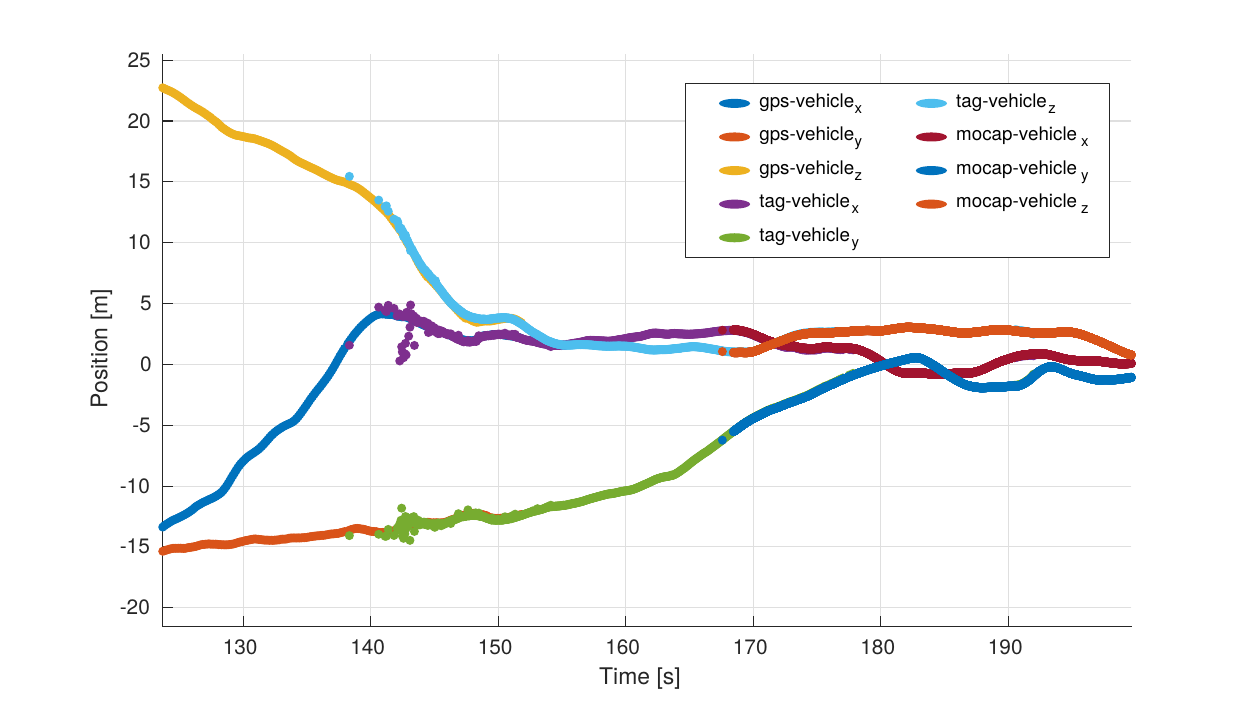}
    \caption{Alignment of the ground~truth segments for the outdoor area, the marker-based transition section, and the indoor motion capture pose. The shown trajectory builds the continuous ground~truth for the outdoor-indoor datasets.}
    \label{fig:transition_segment_alignment}
\end{figure}

As described in the previous sections, the indoor ground~truth based on motion capture data and the fiducial marker-based position of the vehicle are expressed in the reference frame of the motion capture system because the main marker board is a known motion capture object.
Thus, in order to gain continuous ground~truth for the transition datasets, only the outdoor pose information based on the \ac{RTK} \ac{GNSS} and magnetometer solution and the fiducial marker-based vehicle pose need to be aligned.
Since accurate \ac{RTK} \ac{GNSS} solutions become sparse towards the entrance of the building, \ac{GNSS} data from multiple entry approaches are used to increase the accuracy of the alignment. The final alignment between the outdoor \ac{GNSS} information and the fiducial marker-based trajectory is done by using a least-squares solution to Wahba's problem, shown by e.g. \cite{sorkine2017}. The resulting ground~truth after the alignment is shown by Figure \ref{fig:transition_segment_alignment}.
Should this solution be too inaccurate for a specific purpose, the local ground~truth for each segment can be used.

\section{Providing the Data}
\label{sec:providing_the_data}
The data is provided in human-readable plain text format as dedicated CSV files with headers and uncompressed PNG images with attached timestamps in a CSV file.
Individual sensor calibration files are provided in YAML format.
The structure of the files is outlined in Figure~\ref{fig:calib_dir}~and~\ref{fig:main_dir}.
A script for the conversion of this data into ROS bagfiles is provided with the toolbox. The generation of other data formats is possible by using the provided script as a template.
The calibration of all sensors is provided with respect to the PX4~IMU and allows to construct any relative calibration that might be necessary.

Each outdoor dataset provides two ground~truth files. This ground~truth is calculated using the \ac{RTK} \ac{GNSS}~(\SI{8}{\hertz}) and the Magnetometer data~(\SI{80}{\hertz}). Thus, two files are provided, \texttt{ground\_truth\_8hz.csv} provides data calculated based on RAW \ac{GNSS} measurements and time-matched magnetometer measurements, while \texttt{ground\_truth\_80hz.csv} provides data based on interpolated \ac{RTK} \ac{GNSS} measurements and the raw magnetometer data. The usage may depend on the scenario and the choice is left to the user.
Because the ground~truth for the indoor datasets is generated using the motion capture system, \texttt{ground\_truth.csv} provides this data for clear association.

The ground~truth data for the transition dataset contains the segments for outdoor, transition, and indoor poses. The \texttt{ground\_truth.csv} file for these scenarios contains the aligned pose information.
Since data from multiple sources with multiple rates are used, the ground~truth file also has segments with different frame rates.

Further, all data for individual sensor calibration is provided. This includes static IMU data for the generation of the Allan variance, magnetometer intrinsic calibration data, and camera to IMU calibration sequences.

Measurements are not hardware synced, which is the case for most real-world applications and is a strength of the provided dataset.

\section{Usability of the Dataset}
\label{sec:usability_benchmark}
This section provides two bench-marking scenarios that illustrate the datasets usability and how the provided ground~truth relates to the estimated results.
The first scenario uses a state-of-the-art EKF that utilizes multiple sensors and includes a loosely coupled VIO component for an outdoor to indoor transition dataset. The second scenario specifically addresses the VIO navigation aspects by deploying a state-of-the-art VIO framework that solely uses the navigation camera stream and the main IMU.

\subsection{Usage and Result with a state-of-the-art multi-sensor EKF}
\label{sec:usability_transition}
This test uses MaRS, a modular sensor fusion framework introduced by \cite{Brommer2021_mars}, to process outdoor to indoor transition dataset one (see Tab.~\ref{tab:dataset_list}). This represents a possible application in which a continuous state estimate is generated while the sensor data streams are selected/switched based on their availability.

The transition datasets consist of three elements as detailed in Section~\ref{sec:environment_and_experiments}.
The first section begins outdoors, and the filter uses the PX4 IMU, RTK GNSS measurements (position and velocity), vision-based pose as well as one magnetometer.
As the vehicle approaches the building (referring to the map in Fig.~\ref{fig:experiment_map_transition}), Sector~three is entered, and the quality of the GNSS measurements degrade, which triggers a $\chi^2$ rejection test. After the number of rejections reaches a certain threshold, the sensor is considered unreliable and temporarily removed as an input sensor. The estimate continues with VIO pose information and magnetometer.
After passing through the transition phase, the vehicle enters Sector~four by navigating through the entry of the building. This area is prone to magnetic disturbances, and consecutive magnetometer measurement rejections, which recover after the indoor area of the building is entered. At this point, the filter initializes a reference frame for the motion capture measurements, as soon as they become available, to reference the measurements in the navigation world frame.

\begin{figure}[b]
    \centering
    \includegraphics[width=1.0\linewidth, trim=3.5cm 0.4cm 3.2cm 0.9cm, clip]{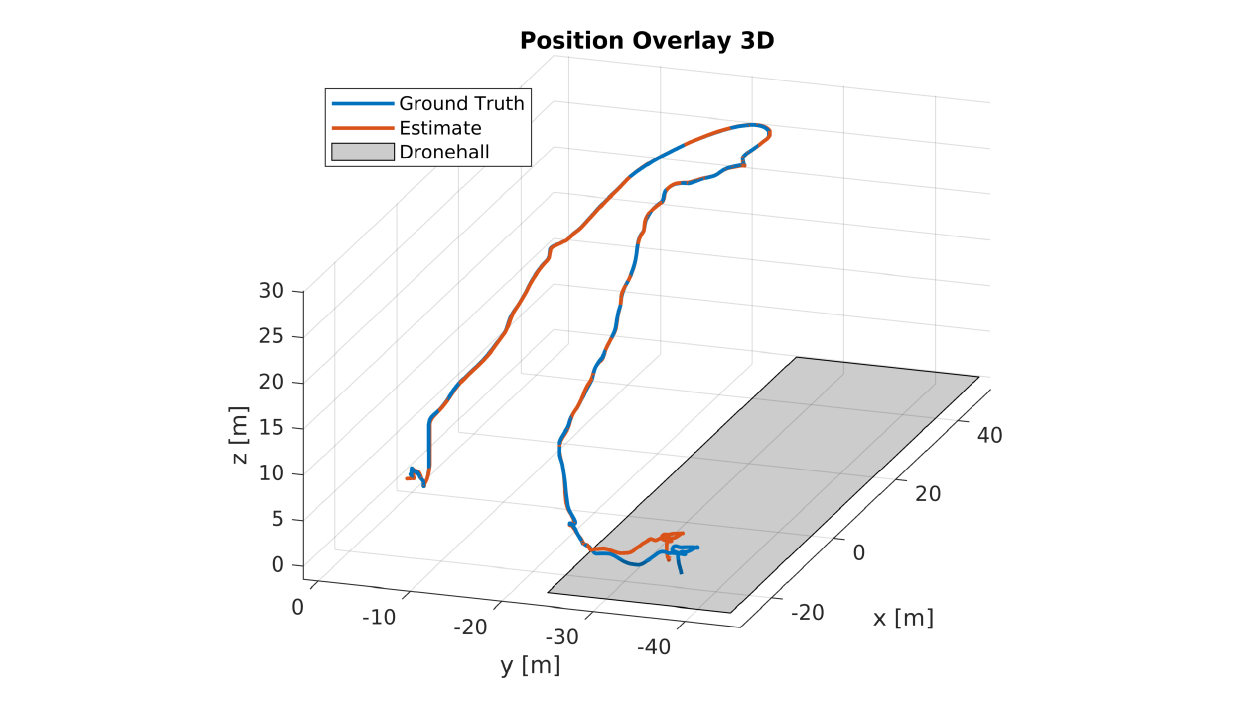}
    \caption{State-of-the-art EKF filtering results for transition dataset one, overlayed with the ground~truth provided by the dataset. The shaded area represents the indoor segment.}
    \label{fig:bench_ekf_result_3d}
\end{figure}

The result for this experiment is shown by Figure~\ref{fig:bench_ekf_result_3d}. As can be seen, the estimate matches the ground~truth data well, despite the transition phase in which sensor switching and consecutive reference frame adaptations occurred. The difference in the indoor section results from a vision drift accumulated during the vision-only phase with challenging vision inputs. The dataset aims to provide the means to improve corresponding algorithms to improve estimation performance for the presented scenarios in the future.

\subsection{Usage and Evaluation of Example Data with State-of-the-Art VIO}
\label{sec:usability_vio}
\begin{figure}[hbt!]
    \centering
    \includegraphics[width=1.0\linewidth, trim=3.2cm 0.8cm 3.0cm 0.8cm, clip]{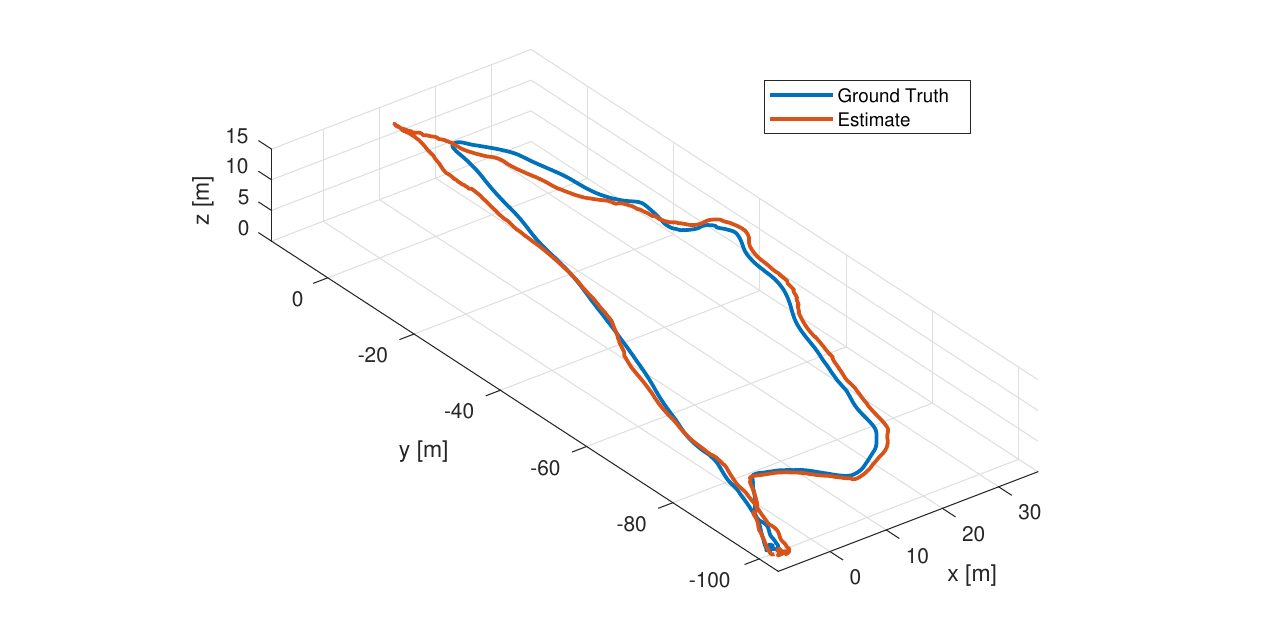}
    \caption{Illustrative visual-inertial localization example using a planar Mars analog dataset with state-of-the-art methods. The estimated trajectory in red in comparison to the provided ground~truth of the dataset (blue). The estimated relative trajectory was globally aligned with the ground~truth.}
    \label{fig:bench_vio_result_3d}
\end{figure}

As a validation of the usability of the datasets for VIO algorithm evaluation, raw IMU data as well as camera images from the Mars analog dataset 18 were used to run OpenVins~\cite{Geneva2020_openvins}, an open-source, state-of-the-art Multi-State Constrained Kalman Filter (MSCKF) based VIO framework.
Figure~\ref{fig:bench_vio_result_3d} shows the estimated trajectory compared with RTK-GNSS based ground~truth. The estimated trajectory and ground~truth are aligned along the four unobservable degrees of freedom of VIO, with the Umeyama alignment algorithm~\cite{Umeyama1991} computed over the full trajectory.
The error of the estimate is calculated after the trajectory alignment and based on the absolute trajectory error (ATE) introduced by \cite{Zhang2018}. The absolute trajectory error of the position is \SI{7.5}{\meter}, and the rotation error is \SI{7}{\degree}.
It is important to note that this dataset poses particular challenges to vision-based algorithms. During the takeoff phase, the shadow of the platform with spinning propellers is in view of the navigation camera, resulting in moving features being tracked even though no motion is experienced by the platform itself.
Moreover, during the flight, an almost featureless ground is seen from the camera, resulting in bad feature tracking (see the top left segment of Fig.~\ref{fig:bench_vio_result_3d}). Novel approaches as introduced by \cite{Hardt-Stremayr2020} are currently tackling this issue.

\section{Lessons Learned}
\label{sec:lessons_learned}
This project posed challenges in terms of system and sensor setup as well as environmental difficulties.
A few aspects have already been outlined in the System Setup Section~\ref{sec:system_setup}. However, this section concludes the main difficulties that had to be overcome.

The generation of ground~truth data posed multiple difficulties. First, the acquisition of highly accurate \ac{GNSS} measurements was not straightforward because the vehicle hosts several computation boards as well as high-frequent data lines, which cause \ac{EMI}.
Thus, the electronic components and the \ac{GNSS} antennas required customized shielding
(see Fig.~\ref{fig:experiment_platform}~and~\ref{fig:rtk_shielding}) to reduce interfering signals to a level at which the \ac{GNSS} provides position accuracy of $1\unit{\centi\meter}$. Section~\ref{sec:rtk_gnss} discusses this issue in detail.
Another difficulty was the generation of continuous ground~truth information for the outdoor to indoor transition area.
This is due to \ac{RTK} \ac{GNSS} signals not being available close to the building area, and the motion capture system not covering the outside of the building entrance.
Thus, a field of fiducial markers was used to provide ground~truth for this transition area. This pose data is aligned with valid \ac{RTK} \ac{GNSS} data for the outdoor area and the motion capture reference frame via marker-object association.
This method allows for global and local ground~truth for each of the three segments of the outdoor-indoor transition trajectories.

Finally, the recording of high amounts of data, as done for this project, was a challenge. Special attention was given to the two onboard embedded platforms to reach balanced computational loads and, most importantly, the full use of available data interfaces. As shown by Figure~\ref{fig:system_diagram}, each board makes use of two storage entities with different data throughputs. USB3 interfaces also needed to be shared between storage devices and sensors. Thus, the data communication and recording needed to be designed to make use of the maximum data capacity that each interface and storage unit can provide. Overall, the system needed to write to an SD and SSD medium on each of the two embedded boards, thus requiring a total of four storage locations to record the data with the given measurement rate.
%

\section{Conclusion and Future Work}
\label{sec:conclusion}
This work introduced the INSANE dataset collection, which aims to provide multi-disciplinary in-flight data with a versatile sensor suite that is subject to real-world sensor effects. The flight scenarios address various research domains for vehicle localization. 
This work discussed individual aspects of the sensors and their integration in detail. The raw data for customized sensor calibrations and the analysis of specific sensor properties such as intrinsic, extrinsic and behavior for vibration behavior in a dedicated test bench setup are provided.

The quality of a dataset is directly correlated with the accuracy and uninterrupted availability of the accompanying ground~truth information. Thus we presented a carefully designed setup to directly measure highly accurate ground~truth as raw data. In contrast to other work, we do not need to filter this data to achieve the high accuracy, which prevents any filter-induced artifacts such as biases and inconsistencies.

Finally, we presented two show-cases which demonstrate the usability of the dataset with comparisons to the provided ground~truth.

The lessons learned throughout this work are a crucial stepping stone for the development and extension of novel flight platforms.
As for future aspects, it is planned to extend the dataset with new scenarios and sensor setups over time.
Thus, the open-sourced dataset is not a static entity and will grow over time, following the same format for compatibility.
Given the versatility of the presented setup, we believe that this data will enable researchers to develop and test future algorithms addressing real-world challenges in various disciplines.
%

\section*{Acknowledgments}
A portion of this research was carried out at the Jet Propulsion Laboratory, California Institute of Technology, under a contract with the National Aeronautics and Space Administration (80NM0018D0004).

The authors would like to thank the Austrian Space Forum (OeWF) for the possibility to perform tests and record datasets during the Analog Mars Mission AMADEE-20 \protect\footnotemark.
\footnotetext{\href{https://oewf.org/amadee-20/}{OeWF - AMADEE20}}

The authors would also like to express a special thanks to the technical personnel, Fred Arneitz, and Patrik Grausberg for their support of the presented work.%

\section*{Funding}
Research was sponsored by the Army Research Office and was accomplished under
Cooperative Agreement Number W911NF-21-2-0245. The views and conclusions contained in this document are
those of the authors and should not be interpreted as representing the official policies, either expressed or implied, of
the Army Research Office or the U.S. Government. The U.S. Government is authorized to reproduce and distribute
reprints for Government purposes notwithstanding any copyright notation herein.

This work has received funding from the European Union’s Horizon 2020 research and innovation programme under grant agreement 871260.

\bibliographystyle{IEEEtran}

\bibliography{bib/christian.bib}

\clearpage
\onecolumn
\appendix

\section{Appendix}
\subsection{Dataset Comparison}
\label{sec:dataset_comparison}
%
%
\begin{table*}[!h]
\setlength\tabcolsep{2pt}
\footnotesize
\vspace{-0.3cm}
\caption{Comparison to other quadcopter datasets}
\label{tab:dataset_comp_copter}

\begin{tabular}[b]{|l|l|>{\raggedright}p{2.1cm}|>{\raggedright}p{2cm}|>{\raggedright}p{2.1cm}|>{\raggedright}p{2cm}|>{\raggedright}p{2cm}|>{\raggedright}p{2cm}|>{\raggedright}p{2cm}|}
\hline 
\multicolumn{2}{|l|}{} & \multicolumn{1}{>{\centering}m{2.1cm}|}{\textbf{INSANE}} & \multicolumn{1}{>{\centering}m{2cm}|}{\textbf{EuRoC \cite{Burri2016}}} & \multicolumn{1}{>{\centering}m{2.1cm}|}{\textbf{UZH-FPV \cite{Delmerico2019}}} & \multicolumn{1}{>{\centering}m{2cm}|}{\textbf{Blackbird \cite{Antonini2020_ijrrblackbird}}} & \multicolumn{1}{>{\centering}m{2cm}|}{\textbf{Upenn Fast Flight \cite{Sun2018}}} & \multicolumn{1}{>{\centering}m{2cm}|}{\textbf{Zurich Urban MAV \cite{Majdik2017}}} & \multicolumn{1}{>{\centering}m{2cm}|}{\textbf{In-flight energy use \cite{Rodrigues2021}}}\tabularnewline
\hline 
\hline 
\multicolumn{2}{|l|}{\textbf{Focus}} & Multi-Sensor fus.

Out-/indoor trans.

Mars Analog & MAV VIO & VIO on Fast Vehicles & Simulated Env. & VIO on fast Vehicles & MAV VIO & Package Delivery\tabularnewline
\hline 
\multicolumn{2}{|l|}{\textbf{Platform}} & Quadcopter & Quadcopter & Quadcopter & Quadcopter & Quadcopter & Quadcopter & Quadcopter\tabularnewline
\hline 
\multicolumn{2}{|l|}{\textbf{Location}} & Indoor, Outdoor,

- Out $\rightarrow$ indoor

- Mars Analog & Indoor & Indoor,

Outdoor & Rendered

environment & Outdoor & Outdoor & Outdoor\tabularnewline
\hline 
\multirow{3}{*}[-1.2cm]{\begin{turn}{90}
\textbf{Camera}
\end{turn}} & \multicolumn{1}{l|}{\textbf{Mono}} & - 1 gray,

~~2056x1542 

~~3MP @20Hz, 

~~global-shutter &  &  &  & - 2 gray, 

~~960x800 

~~0.7MP @40Hz, 

~~global shutter & - 1 RGB, 

~~1920x1080 

~~2MP @30Hz, 

~~rolling shutter & \tabularnewline
\cline{2-9} \cline{3-9} \cline{4-9} \cline{5-9} \cline{6-9} \cline{7-9} \cline{8-9} \cline{9-9} 
 & \textbf{Stereo} & - 1 gray, 

~~848x800 

~~0.6MP @30Hz, 

~~global-shutter & - 1 gray, 

~~768x480 

~~0.3MP @20Hz, 

~~global-shutter & - 1 gray, 

~~640\texttimes 480 

~~0.3MP @50Hz, 

~~global-shutter  &  &  &  & \tabularnewline
\cline{2-9} \cline{3-9} \cline{4-9} \cline{5-9} \cline{6-9} \cline{7-9} \cline{8-9} \cline{9-9} 
 & \textbf{Others} &  &  & - 1 event camera, 

~~346x260 

~~0.09MP @50Hz & - Rendered

~~environment

~~@120Hz &  &  & \tabularnewline
\hline 
\multicolumn{2}{|l|}{\textbf{IMUs}} & - ICM20689

~~@200Hz

- BMI055

~~@200Hz 

- LSM9DS1 

~~@900Hz & - ADIS16448 

~~@200Hz & - miniDAVIS346

~~@1000Hz

- Snapdragon

~~Flight 

~~@500Hz & - Xsens MTi-3

~~@100Hz & - VectorNav 

~~VN-100 

~~@200Hz & - IMU

~~@10Hz & - 3DM-GX5-45 

~~GNSS/INS

~~@10Hz\tabularnewline
\hline 
\multicolumn{2}{|l|}{\textbf{Magnetometer}} & - LSM9DS1 

~~@20Hz

- UST8310 

~~@80Hz &  &  &  &  &  & \tabularnewline
\hline 
\multicolumn{2}{|l|}{\textbf{GNSS}} & - 2 RTK @8Hz

~~3DoF Pos.\&Vel.

- PX4 @5Hz

~~3DoF Position 

~~2DoF Velocity  &  &  &  & - GNSS 

~~@5Hz

~~Pos.\&Vel. &  & - 3DM-GX5-45 

~~GNSS/INS 

~~@10Hz\tabularnewline
\hline 
\multirow{3}{*}[-0.2cm]{\begin{turn}{90}
\textbf{Laser}
\end{turn}} & \textbf{1D} & - Range Finder 

~~@30Hz &  &  &  &  &  & \tabularnewline
\cline{2-9} \cline{3-9} \cline{4-9} \cline{5-9} \cline{6-9} \cline{7-9} \cline{8-9} \cline{9-9} 
 & \textbf{2D} &  &  &  &  &  &  & \tabularnewline
\cline{2-9} \cline{3-9} \cline{4-9} \cline{5-9} \cline{6-9} \cline{7-9} \cline{8-9} \cline{9-9} 
 & \textbf{3D} &  &  &  &  &  &  & \tabularnewline
\hline 
\multicolumn{2}{|l|}{\textbf{UWB}} & - 3 Anchors

~~@7Hz,

~~indoor \& trans. &  &  &  &  &  & \tabularnewline
\hline 
\multicolumn{2}{|l|}{\textbf{Odometry}} & - RealSense

~~T256 V-Slam

~~@200Hz &  &  &  &  &  & \tabularnewline
\hline 
\multicolumn{2}{|l|}{\textbf{Barometer}} & - MS5611

~~@20Hz &  &  &  &  &  & \tabularnewline
\hline 
\multicolumn{2}{|l|}{\textbf{Motion Capture}} & - Optitrack

~~@300Hz & - Vicon

~~@100Hz &  & - Optitrack 

~~@300Hz &  &  & \tabularnewline
\hline 
\multicolumn{2}{|l|}{\textbf{Motor Speed}} & - Setpoint, indoor

~~@100Hz, &  &  & - Encoders 

~~@190Hz &  &  & \tabularnewline
\hline 
\multicolumn{2}{|l|}{\textbf{Others}} & - 125 trans. GT

~~Fiducial marker

~~@20Hz &  &  &  &  &  & Voltage and current sensors\tabularnewline
\hline 
\multicolumn{2}{|l|}{\textbf{GT Method}} & - 6DoF

~~Dual RTK+Mag,

~~8Hz or 50Hz

- Motion Capture

~~@300Hz 

- Fiducial marker 

~~@20Hz & - Leica MS 50

~~3DoF, 20Hz

- Motion Capture 

~~@100Hz & - Leica MS 60 

~~@20Hz & - Motion Capture 

~~@360Hz &  & - Pix4D 

~~visual pose & - 3DM-GX5-45 

~~GNSS/INS 

~~@10Hz\tabularnewline
\hline 
\multicolumn{2}{|l|}{\textbf{Max height}} & 40m & $\S$ & $\S$ & $\S$ & $\S$ & 15m & 100m\tabularnewline
\hline 
\multicolumn{2}{|l|}{\textbf{Max Speed}} & 6m/s & 2.3m/s & 23.4m/s & 7m/s & 17.5m/s & 3.9m/s & 12m/s\tabularnewline
\hline 
\multicolumn{2}{|l|}{\textbf{Reference}} &  & \scriptsize\cite{Schubert2018,Pfrommer2017_penncosyvio} &  &  & \scriptsize\cite{Antonini2020_ijrrblackbird} & \scriptsize\cite{Antonini2020_ijrrblackbird,Schubert2018} & \tabularnewline
\hline 
\multicolumn{9}{l}{$\S$ Values not manually extracted from dataset\vspace{-1cm}}\tabularnewline
\end{tabular}
\vspace{-1cm}
\end{table*}


\clearpage

\begin{table*}
\setlength\tabcolsep{2pt}
\footnotesize 
\caption{Comparison to datasets for outdoor ground robots, such as cars and hand-held robots}
\label{tab:dataset_comp_outdoor}

\begin{tabular}{|c|l|>{\raggedright}p{2.1cm}|>{\raggedright}p{2.2cm}|>{\raggedright}p{2.1cm}|>{\raggedright}p{2.3cm}|>{\raggedright}p{2cm}|>{\raggedright}p{2cm}|>{\raggedright}p{2cm}|}
\hline 
\multicolumn{2}{|l|}{} & \multicolumn{1}{>{\centering}m{2.1cm}|}{\textbf{INSANE}} & \multicolumn{1}{>{\centering}m{2.2cm}|}{\textbf{KITTI}

\textbf{\cite{Geiger2013}}} & \multicolumn{1}{>{\centering}m{2.1cm}|}{\textbf{Malaga Urban} 

\scriptsize\textbf{\cite{Blanco-Claraco2014}}} & \multicolumn{1}{>{\centering}m{2.3cm}|}{\textbf{Oxford RobotCar} \scriptsize\textbf{\cite{Maddern2017_1year1000km}}} & \multicolumn{1}{>{\centering}m{2cm}|}{\textbf{Rosario}

\textbf{\cite{Pire2019_therosariodataset}}} & \multicolumn{1}{>{\centering}m{2cm}|}{\textbf{MADMAX} \scriptsize \textbf{\cite{Meyer2021}}} & \multicolumn{1}{>{\centering}m{2cm}|}{\textbf{NCLT }\scriptsize\textbf{ \cite{Carlevaris-Bianco2016}}}\tabularnewline
\hline 
\hline 
\multicolumn{2}{|l|}{\textbf{Focus}} & Multi-Sensor fus.

Out-/indoor trans.

Mars Analog & Self-Driving Car & Self-Driving Car & Self-Driving Car 

(1 Year) & Agricultural & Mars Analog & Longterm SLAM\tabularnewline
\hline 
\multicolumn{2}{|l|}{\textbf{Platform}} & Quadcopter & Car & Car & Car & Ground robot & - Hand-held

- Ground robot & Segway\tabularnewline
\hline 
\multicolumn{2}{|l|}{\textbf{Location}} & Indoor, Outdoor,

- Out $\rightarrow$ indoor

- Mars Analog & Outdoor & Outdoor &  & Outdoor & Outdoor & Indoor, Outdoor\tabularnewline
\hline 
\multirow{3}{*}[-1.2cm]{\begin{turn}{90}
\textbf{Camera}
\end{turn}} & \multicolumn{1}{l|}{\textbf{Mono}} & - 1 gray,

~~2056x1542 

~~3MP @20Hz, 

~~global-shutter &  &  & - 3 RGB,

~~1024x1024 

~~1MP @11.1Hz 

~~global shutter &  & - 1 RGB 

~~2064x1544,

~~3MP @4Hz & - 6 RGB omni,

~~1600x1200

~~2MP @5Hz\tabularnewline
\cline{2-9} \cline{3-9} \cline{4-9} \cline{5-9} \cline{6-9} \cline{7-9} \cline{8-9} \cline{9-9} 
 & \textbf{Stereo} & - 1 gray, 

~~848x800 

~~0.6MP @30Hz, 

~~global-shutter & - 1 RGB,

~~1392x512 

~~0.7MP @10Hz 

- 1 gray, 

~~1392x512 

~~0.7MP @10Hz & - 1 RGB, 

~~1024x768 

~~0.7MP @100Hz & - 1 RGB, trinocular 

~~1280x960 

~~1.2MP @16Hz & - 1 RGB,

~~672x376 

~~0.2MP @15Hz & - 1 gray,

~~1032x772 

~~0.8MP @14Hz & \tabularnewline
\cline{2-9} \cline{3-9} \cline{4-9} \cline{5-9} \cline{6-9} \cline{7-9} \cline{8-9} \cline{9-9} 
 & \textbf{Others} &  &  &  &  &  & - 2 gray, Omni

~~2064x1544

~~3MP@4-8Hz & \tabularnewline
\hline 
\multicolumn{2}{|l|}{\textbf{IMUs}} & - ICM20689

~~@200Hz

- BMI055

~~@200Hz 

- LSM9DS1 

~~@900Hz & - OXTS RT 3003 

~~@10Hz & - XSensMTi

~~@100Hz &  & - LSM6DS0 

~~@140Hz & - XSENS MTi10 

~~@100Hz & - IMU 

~~@100Hz

- Fiber gyro 

~~@100Hz\tabularnewline
\hline 
\multicolumn{2}{|l|}{\textbf{Magnetometer}} & - LSM9DS1 

~~@20Hz

- UST8310 

~~@80Hz &  &  &  &  &  & \tabularnewline
\hline 
\multicolumn{2}{|l|}{\textbf{GNSS}} & - 2 RTK @8Hz

~~3DoF Pos.\&Vel.

- PX4 @5Hz

~~3DoF Position 

~~2DoF Velocity  & - OXTS RT 3003 

~~@10Hz & - GNSS

~~@1Hz & - NovAtel SPAN

~~CPT ALIGN 

~~IMU+GNSS, 

~~@50 Hz & - 2 RTK 

~~@5Hz & - 2 RTK 

~~@1Hz & - 1 GNSS 

~~@5Hz

- 1 RTK 

~~@1Hz\tabularnewline
\hline 
\multirow{3}{*}[-0.9cm]{\begin{turn}{90}
\textbf{Laser}
\end{turn}} & \textbf{1D} & - Range Finder 

~~@30Hz &  &  &  &  &  & \tabularnewline
\cline{2-9} \cline{3-9} \cline{4-9} \cline{5-9} \cline{6-9} \cline{7-9} \cline{8-9} \cline{9-9} 
 & \textbf{2D} &  &  &  & - 2 SICK

~~LMS-151 &  &  & - Hokuyo @40Hz

- Hokuyo @10Hz\tabularnewline
\cline{2-9} \cline{3-9} \cline{4-9} \cline{5-9} \cline{6-9} \cline{7-9} \cline{8-9} \cline{9-9} 
 & \textbf{3D} &  & - Velodyne 

~~HDL-64E 

~~@10Hz &  & - SICK 

~~LD-MRS &  &  & - Velodyne

~~HDL-32E 

~~@10Hz\tabularnewline
\hline 
\multicolumn{2}{|l|}{\textbf{UWB}} & - 3 Anchors

~~@7Hz,

~~indoor \& trans. &  &  &  &  &  & \tabularnewline
\hline 
\multicolumn{2}{|l|}{\textbf{Odometry}} & - RealSense

~~T256 V-Slam

~~@200Hz &  &  &  & - Wheel encoders &  & \tabularnewline
\hline 
\multicolumn{2}{|l|}{\textbf{Barometer}} & - MS5611

~~@20Hz &  &  &  &  &  & \tabularnewline
\hline 
\multicolumn{2}{|l|}{\textbf{Motion Capture}} & - Optitrack

~~@300Hz &  &  &  &  &  & \tabularnewline
\hline 
\multicolumn{2}{|l|}{\textbf{Motor Speed}} & - Setpoint, indoor

~~@100Hz, &  &  &  &  &  & \tabularnewline
\hline 
\multicolumn{2}{|l|}{\textbf{Others}} & - 125 trans. GT

~~Fiducial marker

~~@20Hz &  &  &  &  &  & \tabularnewline
\hline 
\multicolumn{2}{|l|}{\textbf{GT Method}} & - 6DoF

~~Dual RTK+Mag,

~~8Hz or 50Hz

- Motion Capture

~~@300Hz 

- Fiducial marker 

~~@20Hz & - OXTS RT 3003 

~~(6DoF GPS/IMU) 

~~@10Hz & - GNSS 

~~@1Hz & - NovAtel

~~SPAN-CPT 

~~ALIGN

~~IMU and GNSS

~~@50 Hz &  & - RTK/IMU & - 6DoF 

~~(GPS/IMU/laser)\tabularnewline
\hline 
\multicolumn{2}{|l|}{\textbf{Max height}} & 40m & N/A & N/A & N/A & N/A & N/A & N/A\tabularnewline
\hline 
\multicolumn{2}{|l|}{\textbf{Max Speed}} & 6m/s & $\S$ & $\S$ & $\S$ & $\S$ & 0.48m/s & $\S$\tabularnewline
\hline 
\multicolumn{2}{|l|}{\textbf{Reference}} &  & \scriptsize\cite{Pfrommer2017_penncosyvio},\cite{Schubert2018} & \scriptsize\cite{Schubert2018} &  &  &  & \scriptsize\cite{Pfrommer2017_penncosyvio},\cite{Schubert2018}\tabularnewline
\hline 
\end{tabular}
\vspace{-1cm}
\end{table*}

\clearpage
\begin{table*}
\setlength\tabcolsep{2pt}
\footnotesize 
\caption{Comparison to datasets for indoor ground-robots and hand-held robots}
\label{tab:dataset_comp_indoor}

\begin{tabular}{|c|l|>{\raggedright}p{2.1cm}|>{\raggedright}p{2.25cm}|>{\raggedright}p{2cm}|>{\raggedright}p{2cm}|>{\raggedright}p{2cm}|>{\raggedright}p{2cm}|>{\raggedright}p{2cm}|}
\hline 
\multicolumn{2}{|l|}{} & \multicolumn{1}{>{\centering}m{2.1cm}|}{\textbf{INSANE}} & \multicolumn{1}{>{\centering}m{2.25cm}|}{\textbf{Rawseeds}

\textbf{\scriptsize\cite{Ceriani2009}}} & \multicolumn{1}{>{\centering}m{2cm}|}{\textbf{RGBD-SLAM \scriptsize\cite{Sturm2012}}} & \multicolumn{1}{>{\centering}m{2cm}|}{\textbf{TUM VI \scriptsize\cite{Schubert2018}}} & \multicolumn{1}{>{\centering}m{2cm}|}{\textbf{TUM-LSI}

\textbf{\scriptsize \cite{Walch2017}}} & \multicolumn{1}{>{\centering}m{2cm}|}{\textbf{Naverlabs\scriptsize}

\textbf{\cite{Lee2021}}} & \multicolumn{1}{>{\centering}m{2cm}|}{\textbf{PennCOSYVIO \scriptsize\cite{Pfrommer2017_penncosyvio}}}\tabularnewline
\hline 
\hline 
\multicolumn{2}{|l|}{\textbf{Focus}} & Multi-Sensor fus.

Out-/indoor trans.

Mars Analog & SLAM & RGB-D SLAM & Hand-held VIO & ML camera-based pose est. & Crowded indoor spaces & Hand-held VIO\tabularnewline
\hline 
\multicolumn{2}{|l|}{\textbf{Platform}} & Quadcopter & Ground robot & - Ground robot

- Hand-held & Hand-held & Hand-held & Ground robot & Hand-held\tabularnewline
\hline 
\multicolumn{2}{|l|}{\textbf{Location}} & Indoor, Outdoor,

- Out $\rightarrow$ indoor

- Mars Analog & Indoor, Outdoor & Indoor & Indoor, Outdoor & Indoor, Outdoor & Indoor & Indoor, Outdoor\tabularnewline
\hline 
\multirow{3}{*}[-1.7cm]{\begin{turn}{90}
\textbf{Camera}
\end{turn}} & \multicolumn{1}{l|}{\textbf{Mono}} & - 1 gray,

~~2056x1542 

~~3MP @20Hz, 

~~global-shutter & - 1 RGB, 

~~640x480 

~~0.3MP @30Hz 

- 1 fisheye RGB, 

~~640x640 

~~0.3MP @15Hz & - 1 RGB-D, 

~~640x400 

~~0.3MP @30Hz &  & - 6 RGB,

~~4592\texttimes 3448

~~15MP & - 6 RGB,

~~2592x2048

~~5MP @2.5Hz, 

~~global shutter 

- 4 smartphone 

~~2160x2880 

~~6MP @1Hz, 

~~rolling shutter & - 4 RGB, 

~~1920x1080 

~~2MP @30Hz, 

~~rolling shutter

- fisheye gray,

~~640x480 

~~0.3MP @30Hz\tabularnewline
\cline{2-9} \cline{3-9} \cline{4-9} \cline{5-9} \cline{6-9} \cline{7-9} \cline{8-9} \cline{9-9} 
 & \textbf{Stereo} & - 1 gray, 

~~848x800 

~~0.6MP @30Hz, 

~~global-shutter &  &  & - 1 gray,

~~1024x1024

~~1MP @20Hz &  &  & - 1 gray, 

~~752x480, 

~~0.3MP @20Hz\tabularnewline
\cline{2-9} \cline{3-9} \cline{4-9} \cline{5-9} \cline{6-9} \cline{7-9} \cline{8-9} \cline{9-9} 
 & \textbf{Others} &  & - 1 gray trinocular,

~~640x480 

~~0.3MP @15Hz &  &  &  &  & \tabularnewline
\hline 
\multicolumn{2}{|l|}{\textbf{IMUs}} & - ICM20689

~~@200Hz

- BMI055

~~@200Hz 

- LSM9DS1 

~~@900Hz & @128Hz &  & - BMI160

~~@200Hz &  &  & - 2 acc. 

~~@180Hz

- 2 gyro

~~@100Hz

- 1 acc/gyro

~~@200Hz\tabularnewline
\hline 
\multicolumn{2}{|l|}{\textbf{Magnetometer}} & - LSM9DS1 

~~@20Hz

- UST8310 

~~@80Hz &  &  &  &  &  & \tabularnewline
\hline 
\multicolumn{2}{|l|}{\textbf{GNSS}} & - 2 RTK @8Hz

~~3DoF Pos.\&Vel.

- PX4 @5Hz

~~3DoF Position 

~~2DoF Velocity  & - 1 RTK 

~~@5Hz &  &  &  &  & \tabularnewline
\hline 
\multirow{3}{*}[-1cm]{\begin{turn}{90}
\textbf{Laser}
\end{turn}} & \textbf{1D} & - Range Finder 

~~@30Hz &  &  &  &  &  & \tabularnewline
\cline{2-9} \cline{3-9} \cline{4-9} \cline{5-9} \cline{6-9} \cline{7-9} \cline{8-9} \cline{9-9} 
 & \textbf{2D} &  & - 2 Hokuyo @10Hz

- 2 SICK @75Hz &  &  &  &  & \tabularnewline
\cline{2-9} \cline{3-9} \cline{4-9} \cline{5-9} \cline{6-9} \cline{7-9} \cline{8-9} \cline{9-9} 
 & \textbf{3D} &  &  &  &  &  & - 2 Velodyne 

~~VLP-16 @10Hz & \tabularnewline
\hline 
\multicolumn{2}{|l|}{\textbf{UWB}} & - 3 Anchors

~~@7Hz,

~~indoor \& trans. &  &  &  &  &  & \tabularnewline
\hline 
\multicolumn{2}{|l|}{\textbf{Odometry}} & - RealSense

~~T256 V-Slam

~~@200Hz &  &  &  &  &  & \tabularnewline
\hline 
\multicolumn{2}{|l|}{\textbf{Barometer}} & - MS5611

~~@20Hz &  &  &  &  &  & \tabularnewline
\hline 
\multicolumn{2}{|l|}{\textbf{Motion Capture}} & - Optitrack

~~@300Hz &  &  &  &  &  & \tabularnewline
\hline 
\multicolumn{2}{|l|}{\textbf{Motor Speed}} & - Setpoint, indoor

~~@100Hz, &  &  &  &  &  & \tabularnewline
\hline 
\multicolumn{2}{|l|}{\textbf{Others}} & - 125 trans. GT

~~Fiducial marker

~~@20Hz &  &  &  &  &  & \tabularnewline
\hline 
\multicolumn{2}{|l|}{\textbf{GT Method}} & - 6DoF

~~Dual RTK+Mag,

~~8Hz or 50Hz

- Motion Capture

~~@300Hz 

- Fiducial marker 

~~@20Hz & - 3DoF GPS

- 2D+heading

~~(visual tags/laser) & - Motion capture & - Motion capture 

~~@120Hz

~~(partial) & - NavVis M3

~~Lidar based 

~~pose & - LiDAR SLAM 

- SFM & - 6DoF

~~Fiducial marker\tabularnewline
\hline 
\multicolumn{2}{|l|}{\textbf{Max height}} & 40m & N/A & N/A & N/A & N/A & N/A & N/A\tabularnewline
\hline 
\multicolumn{2}{|l|}{\textbf{Max Speed}} & 6m/s & $\S$ & $\S$ & $\S$ & $\S$ & $\S$ & $\S$\tabularnewline
\hline 
\multicolumn{2}{|l|}{\textbf{Reference}} &  & \scriptsize\cite{Pfrommer2017_penncosyvio} & \scriptsize\cite{Pfrommer2017_penncosyvio} & \scriptsize\cite{Schubert2018} &  &  & \scriptsize\cite{Schubert2018}\tabularnewline
\hline 
\end{tabular}
\vspace{-1cm}
\end{table*}


\twocolumn
\subsection{File Structure}
\begin{figure}[hbt!]
    \centering
\begin{forest}
  pic dir tree,
  pic root,
  where level=0{}{
    directory,
  },
  [calibrations
      [magnetometer\_and\_imu(raw)
        [<location>
            [<cal\_method>
                [lsm\_imu.csv, file]
                [lsm\_mag.csv, file]
                [px4\_baro.csv, file]
                [px4\_imu.csv, file]
                [px4\_mag.csv, file]
            ]
        ]
    ]
    [sensors\_and\_cameras(processed)
        [camera\_calibration\_<location>.yaml, file]
        [mag\_calibration\_<location>.yaml, file]
        [sensor\_calibration.yaml, file]
    ]
    [static\_sensor\_data(raw)
        [lrf\_range.csv, file]
        [lsm\_imu.csv, file]
        [lsm\_mag.csv, file]
        [px4\_baro.csv, file]
        [px4\_imu.csv, file]
        [px4\_mag.csv, file]
        [rs\_imu.csv, file]
        [uwb\_range.csv, file]
    ]
    [<nav/stereo>\_camera(raw)
        [intrinsic]
        [extrinsic]
    ]
  ]
\end{forest}
    \caption{Calibration directory}
    \label{fig:calib_dir}
\end{figure}
\begin{figure}[hbt!]
    \centering
\begin{forest}
  pic dir tree,
  pic root,
  where level=0{}{
    directory,
  },
[$<$xyz$>$\_experiment\_data
  [ground\_truth
    [ground\_truth\_8hz.csv, file]
    [ground\_truth\_80hz.csv, file]
  ]
  [nav\_cam
    [nav\_cam\_timestamps.csv, file]
    [img 
       [$<$xyz$>$.png, file]
    ]
  ]
  [stereo\_cam
    [stereo\_cam\_timestamps.csv, file]
    [img
      [$<$xyz$>$\_c1.png, file]
      [$<$xyz$>$\_c2.png, file]
    ]
  ]
  [lrf\_range.csv, file]
  [lsm\_imu.csv, file]
  [lsm\_mag.csv, file]
  [px4\_baro.csv, file]
  [px4\_gps.csv, file]
  [px4\_gps\_vel\_data.csv, file]
  [px4\_imu.csv, file]
  [px4\_mag.csv, file]
  [px4\_rpm.csv, file]
  [rs\_imu.csv, file]
  [rs\_odom.csv, file]
  [rtk\_gps1.csv, file]
  [rtk\_gps2.csv, file]
  [mocap\_tag\_board.csv, file]
  [mocap\_vehicle\_data.csv, file]
  [tags.csv, file]
  [uwb\_range.csv, file]
  [time\_info.yaml, file]
]
\end{forest}
    \caption{Main data directory}
    \label{fig:main_dir}
    \vspace{-1cm}
\end{figure}

\end{document}